\documentclass[lettersize,journal,onecolumn]{IEEEtran}

\usepackage{setspace}
\usepackage{amsmath,amsfonts}

\usepackage{array}
\usepackage{caption}
\usepackage{textcomp}
\usepackage{stfloats}
\usepackage{url}
\usepackage{verbatim}
\usepackage{tabularx}
\usepackage[T1]{fontenc}
\usepackage{graphicx}
\usepackage{cite}
\usepackage{float}
\usepackage{longtable}
\usepackage{subfloat}
\usepackage{multirow}
\usepackage{soul}
\usepackage[table]{xcolor}
\usepackage{epstopdf}
\allowdisplaybreaks[0]
\usepackage{subfigure}
\usepackage[switch]{lineno}
\makeatletter
\newif\if@restonecol
\makeatother

\usepackage[linesnumbered,ruled,vlined]{algorithm2e}
\usepackage{algpseudocode}
\usepackage{amsmath}

\begin{document}
\title{Optimizing Age of Information in Vehicular Edge Computing with Federated Graph Neural Network Multi-Agent Reinforcement Learning}

\author{

{
	Wenhua Wang, 
	Qiong Wu,~\IEEEmembership{Senior Member,~IEEE}, 
	Pingyi Fan,~\IEEEmembership{Senior Member,~IEEE},\\ 
	Nan Cheng, ~\IEEEmembership{Senior Member,~IEEE}, 
	Wen Chen, ~\IEEEmembership{Senior Member,~IEEE},\\ 
	Jiangzhou Wang, ~\IEEEmembership{Fellow,~IEEE} 
	and Khaled B. Letaief, ~\IEEEmembership{Fellow,~IEEE}
}

\thanks{

{	Qiong Wu and Wenhua Wang are with the School of Internet of Things Engineering, Jiangnan University, Wuxi 214122, China, (e-mail: qiongwu@jiangnan.edu.cn, wenhuawang@stu.jiangnan.edu.cn)

	Pingyi Fan is with the Department of Electronic Engineering, Beijing National Research Center for Information Science and Technology, Tsinghua University, Beijing 100084, China (Email: fpy@tsinghua.edu.cn)
	
	Nan Cheng is with the State Key Lab. of ISN and School of Telecommunications Engineering, Xidian University, Xi'an 710071, China (e-mail:
		dr.nan.cheng@iece.org)
		
	Wen Chen is with the Department of Electronic Engineering, Shanghai Jiao Tong University, Shanghai 200240, China(e-mail: wenchen@sjtu.edu.cn)

	Jiangzhou Wang is with the School of Engineering, University of Kent, CT2 7NT Canterbury, U.K. (Email: j.z.wang@kent.ac.uk)

	K. B. Letaief is with the Department of Electrical and Computer Engineering, the Hong Kong University of Science and Technology (HKUST), Hong Kong(e-mail:eekhaled@ust.hk)
	
	%
%
}

}
}



\maketitle
\begin{abstract}
With the rapid development of intelligent vehicles and Intelligent Transport Systems (ITS), the sensors such as cameras and LiDAR installed on intelligent vehicles provides higher capacity of executing computation-intensive and delay-sensitive tasks, thereby raising deployment costs. To address this issue, Vehicular Edge Computing (VEC) has been proposed to process data through Road Side Units (RSUs) to support real-time applications. This paper focuses on the Age of Information (AoI) as a key metric for data freshness and explores task offloading issues for vehicles under RSU communication resource constraints. We adopt a Multi-agent Deep Reinforcement Learning (MADRL) approach, allowing vehicles to autonomously make optimal data offloading decisions. However, MADRL poses risks of vehicle information leakage during communication learning and centralized training. To mitigate this, we employ a Federated Learning (FL) framework that shares model parameters instead of raw data to protect the privacy of vehicle users. Building on this, we propose an innovative distributed federated learning framework combining Graph Neural Networks (GNN), named Federated Graph Neural Network Multi-Agent Reinforcement Learning (FGNN-MADRL), to optimize AoI across the system. For the first time, road scenarios are constructed as graph data structures, and a GNN-based federated learning framework is proposed, effectively combining distributed and centralized federated aggregation. Furthermore, we propose a new  MADRL algorithm that simplifies decision making and enhances offloading efficiency, further reducing the decision complexity. Simulation results demonstrate the superiority of our proposed approach to other methods through simulations.
\end{abstract}

\section{Introduction}

\IEEEPARstart{W}{ith} autonomous driving technology and Intelligent Transport Systems (ITS) evolving, intelligent vehicles  deploy sensors like cameras and LiDAR to aid driving or achieve automation\cite{ref1,ref32,ref33,ref44}. These technologies heighten demands for computational and storage resources, thereby posing challenges for resource-limited vehicles\cite{ref2,ref34,ref35,ref45}.
Vehicular Edge Computing (VEC) is a promising approach to support real-time applications by enabling vehicles to offload tasks to nearby Road Side Units (RSUs), which process these tasks with their substantial computation and storage capabilities and then return the results to the vehicles\cite{ref3,ref4,ref36,ref46}.
Moreover data freshness is becoming increasingly important in VEC. Different from traditional performance metrics, the Age of Information (AoI) is a key indicator of data freshness, considering the generation time and transmission delay of data \cite{ref1,ref5,ref47}. 
However, as the number of vehicles or computation tasks increases, transmission interference between vehicles will greatly increase, potentially deteriorating the AoI for each vehicle's task\cite{ref6,ref37,ref38}. Additionally, relying on RSUs for  vehicle task offloading will increase the information transmission overhead between vehicles and RSUs, and it will also degrade the AoI requirements. Therefore, distributed collaborative offloading among vehicles is crucial for optimizing the AoI in VEC.
 
In recent years, Multi-agent Deep Reinforcement Learning (MADRL) has offered a new solution for multi-vehicle task offloading\cite{ref7,ref8,}. Each vehicle acts as an individual agent, making optimal data offloading decisions based on its observations, enabling decentralized decision-making without waiting for centralized scheduling.
 However, most MADRL training currently relies on the learning model exchange communication and centralized training, where vehicles' training and decision-making require other vehicles' decision and state information, increasing communication bandwidth resource consumption\cite{ref9,ref10}. Additionally, when RSUs collect all vehicle data for training, it will face the high risk of vehicle information leakage. This leads to distrust of RSUs and increases the risk of intercepting raw data, creating major bottlenecks for training  models\cite{ref11,ref39,ref41}.

Federated Learning (FL) offers a potential solution, where vehicles can collaboratively train models without sharing raw data. This is achieved by sending model or gradients instead of raw data to RSUs, thus protecting vehicle users' data\cite{ref12,ref13}. 
By aggregating models from various vehicles, FL facilitates the sharing of knowledge and learning experiences, resulting in a more comprehensive and accurate MADRL model.
However, traditional FL uses average aggregation, treating each vehicle's contribution equally, neglecting that each vehicle may have different data features and varying contributions to model training\cite{ref14,ref42,ref43}. 
In vehicular scenarios, the mobility of vehicles generates rich topological information, and each vehicle possesses unique features, such as speed, data, and the quality of model training. These personalized details can effectively enhance the generalization capability of MADRL models. 
Graph Neural Networks (GNNs), widely applied in various domains for their ability to extract graph information\cite{ref15}, effectively capture features from vehicle-road graph and learn from the graph to generate FL aggregated model weights.
Currently, there is no research considering the enhancement of FL training for MADRL models using vehicle-road graphs to reduce AoI. This gap is the motivation for our work.

In this article, we introduce a novel Federated Graph Neural Network Multi-Agent Reinforcement Learning (FGNN-MADRL) algorithm, aimed at optimizing the AoI in VEC. Our approach incorporates vehicle-road graph with distributed FL to enhance the training of MADRL models. The major contributions of this paper are summarized as follows\footnote{The source code has been released at: https://github.com/qiongwu86/Optimizing-AoI-in-VEC-with-Federated-Graph-Neural-Network-Multi-Agent-Reinforcement-Learning}:

\begin{itemize}
	\item{This paper introduces a method to construct vehicle scenes as vehicle-road graph for the first time. Specifically, roads are divided into segments, each considered a node in a GNN, and establishing edges based on vehicle-to-vehicle communication. This innovative approach effectively addresses the challenges posed by the dynamic variation in vehicle number.}
	
	\item{We propose an innovative GNN-based distributed FL framework that combines distributed local federated aggregation with centralized global federated aggregation. The distributed local federated aggregation, informed by GNN-extracted vehicular road graph structure, effectively generates weights for federated aggregation, considering each vehicle's unique features and contributions. The centralized global federated aggregation further enhances overall model stability and comprehensive capability by integrating all local models.}
	
	\item{Additionally, we present a new MADRL algorithm for efficient cooperative offloading among vehicles. In this algorithm, each vehicle makes decisions based solely on its observations, independent of other vehicles' decisions and observations. This substantially simplifies decision-making and enhances offloading efficiency. By reducing reliance on external information, this MADRL algorithm effectively improves the adaptability and reliability of vehicle offloading strategies.}
\end{itemize}

The remainder of this paper is organized as follows. Section II presents related work. Section III describes our system model, which includes the system scenario, communication model, AoI model and the problem we aim to address. In Section IV, we propose the our FGNN-MADRL scheme. This section begins with an introduction to the GNN-based FL algorithm, followed by an explanation of the GNN combined multi-agent SAC framework and algorithm. Section V is dedicated to simulation experiments and analysis. Finally, we conclude it in Section VI.

\section{Related Work}
\label{Related Work}
In this part, we first review the research on GNN in VEC, followed by an overview of cooperative task offloading.
\subsection{Application of GNN in VEC}

Recent studies have begun to apply GNNs in IoV. 
Liu \emph{et al.} in \cite{ref1} proposed a Spatio-temporal Modeling And ReconsTruction (SMART) framework for assessing the feasibility of different time-delay sensitive services in large-scale IoV. This framework models the VANET as a graph by dividing the service area into subareas (nodes) connected by edges representing similar delay probabilities. SMART leverages Graph Convolutional Networks (GCN) and Deep Q-Networks (DQN) to capture spatial and temporal features of the graph, enabling the reconstruction of an updated large-scale VANET topology from limited subarea samples.
He \emph{et al.} in \cite{ref17} proposed a distributed spectrum sharing framework enhanced by GNN for vehicular networks. They represented the vehicular network as a graph with local observations of vehicle pairs as nodes and channel gains of interference links as edges. The GNN learns low-dimensional features of each node/vehicle pair, with each pair treated as an agent in MADRL. This approach optimizes the total capacity of the vehicular network and base station links using MADRL, with information propagated along graph edges to update each vehicle without base station support.
In \cite{ref18}, He \emph{et al.} addressed the challenge of task allocation in multi-scale IoV with a Deep Reinforcement Learning-based efficient task allocation scheme. They combined storage, computation, and caching mechanisms to support vehicular task distribution across multiple system scales. The dynamic system was modeled graphically, incorporating node characteristics and time IoV-varying edge relationships, using a Graph Attention Network (GAT) and a hybrid algorithm combining Deep Deterministic Policy Gradient (DDPG) for task scheduling optimization.
Zhou \emph{et al.} in \cite{ref19} presented a computational task allocation method with demand prediction and RL for Internet of Things (IoT) environments supported by 6G technology. They used a spatial-temporal GNN-based prediction method for task demand forecasting and a simplex algorithm for cache decision-making. Additionally, they proposed a Twin Delayed DDPG (TD3)-based computational task allocation method.
Liu \emph{et al.} in \cite{ref20} proposed a GNN and DRL-based GA-DRL algorithm for the subtask-to-vehicle assignment problem in Directed Acyclic Graph (DAG) tasks within IoV. They used multi-head GAT networks to extract subtask feature information, integrating these features into Double Deep Q-Networks (DDQN) for decision-making.
Xiao \emph{et al.} in \cite{ref21} introduced a Stochastic Graph Neural Network (SGNN) and RL-based distributed stochastic decision algorithm for intelligent traffic control tasks. It tried to capture dynamic topological connectivity features of vehicles using SGNN, enabling disturbance resistance, where the SGNN is embedded in a Proximal Policy Optimization (PPO) framework with a value decomposition function for modeling vehicle relationships as random graphs.
Chen \emph{et al.} in \cite{ref22} proposed a resource orchestration algorithm for vehicular cloud computing networks, abstracting the resource orchestration problem into a virtual network embedding problem. They designed a four-layer policy network based on GCN to calculate node embedding probabilities, extracting spatial structural information between nodes and neighborhoods.
Li \emph{et al.} in \cite{ref23} developed a GCN-based topology design method (G-DFL) to improve training efficiency in VANET distributed federated learning. They extracted wireless network topology features between vehicles using GCN, optimizing training delays to generate connection graphs. Moreover, they used the Christofides algorithm to find a minimum delay Hamiltonian circuit for model sharing.
However, these studies did not consider the dynamic nature of vehicles, such as changes in network node topology due to vehicle movement.
\subsection{Cooperative Offloading in VEC}
Recent research has explored collaborative task offloading in vehicular networks.
In \cite{ref24}, Lang \emph{et al.}  proposed a blockchain-based data sharing architecture, targeting information sharing and computational offloading in vehicular multi-access edge computing (MEC) networks. This architecture, using blockchain technology, aims to provide accurate service vehicle information to support cooperative computation offloading. To facilitate effective decision-making and data synchronization, the authors introduced a consensus mechanism combining service proof and practical Byzantine fault tolerance, along with a game theory-based offloading decision model, designed to guide user vehicles in making appropriate choices in cooperative computation offloading scenarios. 
In \cite{ref26}, M. Zaki \emph{et al.}  introduced a Cooperative Perception-based Task Offloading (CPTO) scheme, aimed at optimizing vehicular edge computing (VEC) in autonomous vehicles. CPTO focuses on maximizing vehicles' cooperative perception capabilities and minimizing the latency of perception aggregation, adhering to specific deadlines. To achieve this, they formulated the task offloading problem as a multi-objective 0-1 integer linear programming (0-1 ILP) and proposed a greedy heuristic algorithm, CPTO-Heuristic (CPTO-H), to solve this optimization problem. 
In \cite{ref27}, Wang \emph{et al.}  proposed a novel multi-user computational offloading game method for vehicular MEC networks, adjusting the offloading probability of each vehicle. This method considers the distance between vehicles and MEC access points, application and communication models, and the competitive scenario of multiple vehicles for MEC resources, then designed a payoff function. Additionally, the authors built a distributed optimal response algorithm based on the computational offloading game model, aiming to maximize the utility of each vehicle. 
In \cite{ref28}, Zhou \emph{et al.}  proposed a method to optimize offloading decision thresholds in MEC networks, intending to maximize the expected successful offloading rate of tasks. Combining game theory analysis and constrained nonlinear optimization theory, they showed that at least one mixed-strategy Nash equilibrium exists in the system. They formulated task offloading optimization as a Multi-Agent decision problem and developed a distributed unconstrained Lagrangian optimization (ULO) scheme based on the best response mechanism. 
In \cite{ref29}, Alam \emph{et al.}  introduced an innovative three-tier vehicular-assisted multi-access edge computing (VMEC) network design to address collaborative computational offloading issues in high-mobility vehicular network environments. This network utilizes moving and parked vehicles associated with RSUs as VMEC servers in the fog layer and proposed a strategy based on the Hungarian algorithm in Multi-Agent to find the optimal offloading strategy through collaborative agents' actions in dynamic network environments. 
In \cite{ref30}, He \emph{et al.}  researched dynamic data offloading in urban rail transit, aiming to improve the low latency and stability issues of vehicle-to-ground communication caused by high-speed mobility. Using a software-defined network (SDN) controller, they enabled mobile users to choose MEC servers for offloading their data. To decide the specific MEC servers for mobile users' data offloading, the authors conducted a game among mobile users and formulated an optimization problem of the user utility function to determine the optimal offloading data volume for MEC servers and maximize user utility. However, none of the above studies considered the topological structure in vehicular scenarios.

As mentioned, no work has considered collaborative task offloading concerning the dynamic topological structure of vehicles.

\section{System Model}

In this section, we will introduce the system model. We first describe the system scenario, followed by the communication model. Next, we introduce the AoI model and finally present the problem that needs to be addressed.
\begin{figure}[ht]
	\center
	\includegraphics[scale=0.35]{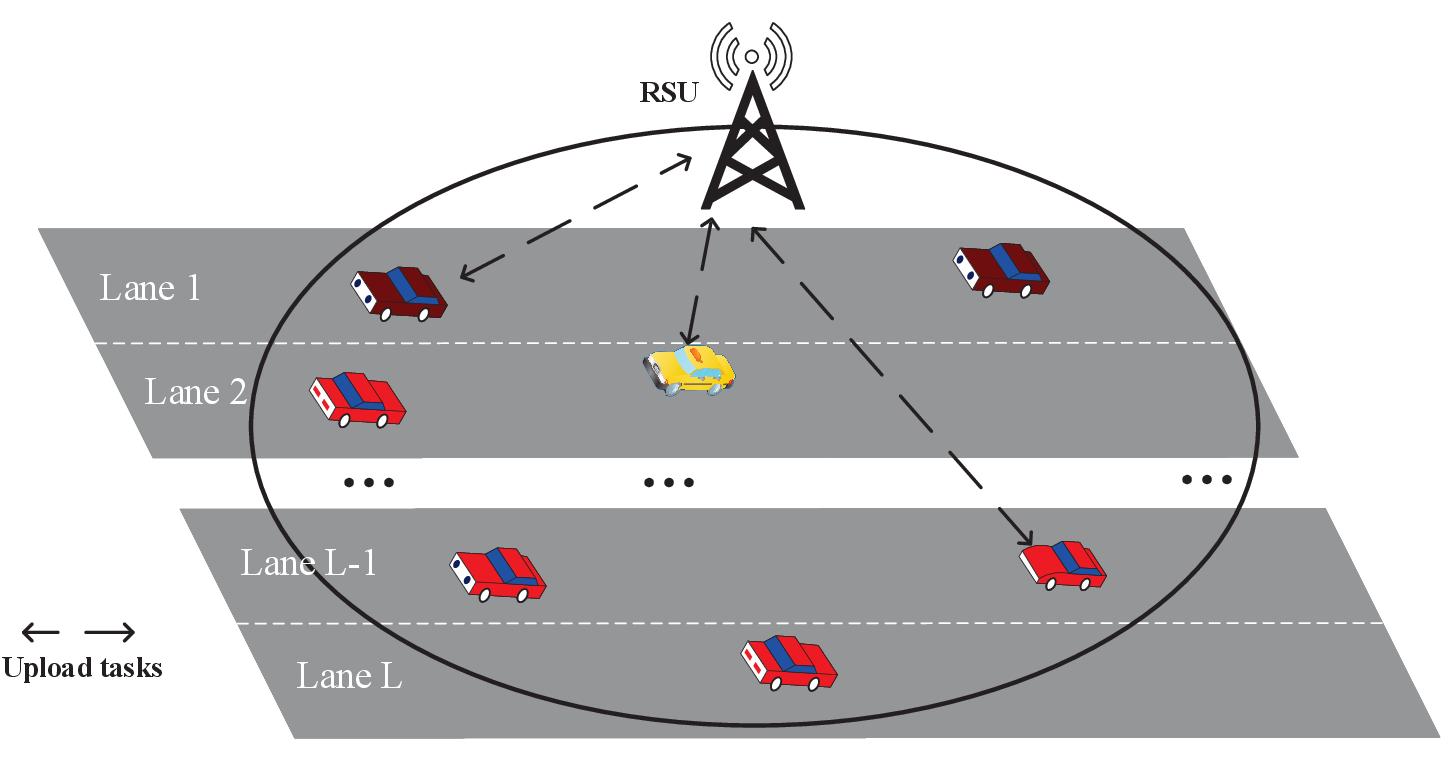}
	\caption{System Scenario}
	\label{fig_system_model}
\end{figure}

\subsection{System Scenario}
As depicted in Fig. \ref{fig_system_model}, consider a VEC scenario where RSUs are deployed alongside the road, each with a communication coverage radius of ${{D}_{r}}$. The VEC scenario contains $\mathcal{L}=\{1,2,3,...,L\}$ lanes, where $L$ represents the number of lanes. Each lane limits the vehicles to different speeds, denoted by $\mathcal{V}=\{v^{1},v^{2},...,v^{i},...,v^{L}\}$. We assume uniform speed for vehicles on each lane. Vehicle generation on each lane follows a Poisson distribution $\mathcal{\lambda} =\{{{\lambda }^{1}},{{\lambda }^{2}},...,{{\lambda }^{i}},...,{{\lambda }^{L}}\}$, where ${{\lambda }^{i}}$ represents the arrival rate of vehicles entering the RSU coverage area on the $i^{th}$ lane. We denote the set of all vehicles within the RSU coverage area at time slot $t$ as ${{\mathbb{V}}_{t}}$. Each vehicle randomly generates a task, with the task generation interval following a Poisson distribution with parameter $\mu$. The size of each task follows a uniform distribution within the range $[{{d}_{\min }},{{d}_{\max }}]$, where ${{d}_{\min }}$ and ${{d}_{\max }}$ represent the minimum and maximum task sizes, respectively. Once a task is generated, each vehicle needs to transmit the task to the RSU for processing. The tasks will first be stored in a queue awaiting transmission, and the tasks in the queue will be sent according to the First-in-First-Out (FIFO) policy. 

We use ${{J}_{{{c}_{i}}}}(t)=\{1,2,3,...,{{J}_{{{c}_{i,\max}}}}(t)\}$ to represent the task index in the queue of vehicle ${{c}_{i}}$ at time slot $t$, where ${{J}_{{{c}_{i,\max}}}}(t)$ indicates the maximum index of tasks for the current time slot, which also indicates the number of tasks in the queue. Specifically, ${{J}_{{{c}_{i}}}}(t)=1$ represents the earliest generated task in the queue, i.e., the current task awaiting transmission.
If a new task is generated, the number of tasks in the queue increases by one, i.e., ${{J}_{{{c}_{i}}}}(t)=\{1,2,3,...,{{J}_{{{c}_{i,\max}}}}(t)\}$.
\subsection{Communication Model}
Assuming the transmitting power of vehicle ${{{c}}_{i}}$ at time slot $t$ is ${{p}_{{{c}_{i}}}}(t)$. Thus the transmission rate of vehicle ${{{c}}_{i}}$ at time slot $t$ can be calculated based on Shannon's theorem, i.e.,
\begin{equation}
{{r}_{{{{c}}_{i}}}}(t)=B{{\log }_{2}}\left( 1+\frac{{{g}_{{{c}_{i}}}}\left( t \right){{p}_{{{c}_{i}}}}\left( t \right)}{\sum\nolimits_{{{c}_{i}}\ne {{c}_{j}}}{{{g}_{{{c}_{j}}}}\left( t \right){{p}_{{{c}_{j}}}}\left( t \right)+{{\sigma }^{2}}}} \right),
\label{eq1}
\end{equation}	 
where $B$ is the total uplink bandwidth within the RSU coverage area, ${{g}_{{{c}_{i}}}}(t)$ is the channel gain and ${{\sigma }^{2}}$ is the noise power. ${{p}_{{{c}_{j}}}}(t)$ and ${{g}_{{{c}_{j}}}}(t)$ are the transmitting power and channel gain of other vehicles, respectively. ${{g}_{{{c}_{i}}}}(t)$ is calculated as
\begin{equation}
{{g}_{{{c}_{i}}}}(t)=\sqrt{{{\alpha }_{{{c}_{i}}}}(t)}{{h}_{{{c}_{i}}}}(t),
\label{eq2}
\end{equation}
where ${{\alpha }_{{{c}_{i}}}}(t)$ and ${{h}_{{{c}_{i}}}}(t)$ represent the large-scale and small-scale fading components of vehicle ${{{c}}_{i}}$ at time slot $t$, respectively. The large-scale component ${{\alpha }_{{{c}_{i}}}}(t)$ includes path loss and log-normal shadowing. Let ${{X}_{r}}=({{x}_{r}},{{y}_{r}})$ be the RSU's coordinate, and the vehicle ${{{c}}_{i}}$'s coordinatesat time slot $t$ is ${{X}_{{{c}_{i}}}}(t)=({{x}_{{{c}_{i}}}}(t),{{y}_{{{c}_{i}}}}(t))$. The large-scale fading component ${{\alpha }_{{{c}_{i}}}}(t)$ can be calculated as
\begin{equation}
{{\alpha }_{{{c}_{i}}}}(t)={PL}({{X}_{{{c}_{i}}}}(t),{{X}_{r}})+{{\chi }_{{{c}_{i}}}}(t),
\label{eq3}
\end{equation}
where ${PL}\left( \cdot  \right)$ indicates distance-related path loss. ${{\chi }_{{{c}_{i}}}}(t)$ represents the log-normal shadow fading from ${{X}_{{{c}_{i}}}}(t)$ to ${{X}_{r}}$, which is updated as
\begin{equation}
{{\chi }_{{{c}_{i}}}}(t)=\rho _{{{c}_{i}}}^{1}(t){{\chi }_{{{c}_{i}}}}(t-1)+{{\sigma }_{s}}{{e}_{{{c}_{i}}}}(t),
\label{eq4}
\end{equation}
where  ${{e}_{{{c}_{i}}}}(t)$ is a Gaussian-distributed log-normal shadow fading random variable.
$\rho _{{{c}_{i}}}^{1}(t)$ is the correlation coefficient of shadow fading and it is calculated as
\begin{equation}
\rho _{{{c}_{i}}}^{1}(t)={{e}^{\frac{\Delta ({{X}_{{{c}_{i}}}}(t))}{{{d}_{{cor}}}}}},
\label{eq5}
\end{equation}
where $\Delta ({{X}_{{{c}_{i}}}}(t))=||{{X}_{{{c}_{i}}}}(t)-{{X}_{{{c}_{i}}}}(t-1)|{{|}_{2}}$ indicates the Euclidean distance between vehicle ${{{c}}_{i}}$ at time slots $t$ and $t-1$, ${{d}_{{cor}}}$ is the correlation length of the environment.

We adopt the Jakes fading model to introduce the small-scale Rayleigh fading component ${{h}_{{{c}_{i}}}}(t)$ as a first-order complex Gaussian Markov process, i.e.,
\begin{equation}
{{h}_{{{c}_{i}}}}(t)=\rho _{{{c}_{i}}}^{2}(t)\cdot {{h}_{{{c}_{i}}}}(t-1)+{{q}_{{{c}_{i}}}}(t),
\label{eq6}
\end{equation}
where $\rho _{{{c}_{i}}}^{2}(t)$ represents the correlation coefficient between ${{h}_{{{c}_{i}}}}(t)$ and ${{h}_{{{c}_{i}}}}(t-1)$, and ${{q}_{{{c}_{i}}}}(t)$ is an independent channel innovation process. The correlation coefficient $\rho _{{{c}_{i}}}^{2}(t)$ is calculated as $\rho _{{{c}_{i}}}^{2}(t)={{J}_{0}}(2\pi {{f}_{d,{{c}_{i}}}}(t)\tau )$, where ${{J}_{0}}(\cdot )$ is the zeroth-order Bessel function of the first kind and $\tau $ is the time length of a time slot. ${{f}_{d,{{c}_{i}}}}(t)=\frac{{{v}_{{{c}_{i}}}}(t)\cdot {{f}_{c}}}{c}$ is the Doppler frequency considering the impact of vehicle movement, where ${{v}_{{{c}_{i}}}}(t)$ is the speed of the vehicle $c_i$, ${{f}_{c}}$ is the carrier frequency ${{f}_{c}}$, as ${{f}_{d,{{c}_{i}}}}(t)=\frac{{{v}_{{{c}_{i}}}}(t)\cdot {{f}_{c}}}{c}$, where $c=3\times {{10}^{8}}$ is the speed of light. The independent channel innovation process ${{q}_{{{c}_{i}}}}(0),{{q}_{{{c}_{i}}}}(1),{{q}_{{{c}_{i}}}}(2),\ldots$ consists of independently distributed circularly symmetric complex Gaussian (CSCG) random variables, with distribution $C\mathcal{N}\left( 0,1-{{(\rho _{{{c}_{i}}}^{2}(t))}^{2}} \right)$. It is important to note that the initial small-scale Rayleigh fading component ${{h}_{{{c}_{i}}}}(0)$ follows a $C\mathcal{N}\left( 0,1 \right)$ distribution.
\subsection{AoI Model}
We use ${{\phi }_{{{{c}}_{i}},{{J}_{{{{c}}_{i}}}}}}(t)$ to represent the AoI of the task ${{J}_{{{{c}}_{i}}}}(t)$ of vehicle ${{{c}}_{i}}$ at time slot $t$. The AoI of the ${{J}_{{{{c}}_{i}}}}(t)=1$ task, denoted as ${{\phi }_{{{{c}}_{i}},1}}(t)$, is calculated as
\begin{equation}
{{\phi }_{{{{c}}_{i}},1}}\left( t \right)=\left\{ \begin{matrix}
   {{\phi }_{{{{c}}_{i}},1}}\left( t-1 \right)+\frac{{{d}_{{{{c}}_{i}},1}}\left( t \right)}{{{r}_{{{{c}}_{i}}}}(t)},{if } \quad { {r}_{{{{c}}_{i}}}}(t)\cdot \tau \ge {{d}_{{{{c}}_{i}},1}}
    \\
   {{\phi }_{{{{c}}_{i}},1}}\left( t-1 \right)+\tau ,{ otherwise                             }  \\
\end{matrix} \right.
\label{eq7},
\end{equation}
where ${{d}_{{{{c}}_{i}},1}}\left( t \right)$ represents the size of the ${{J}_{{{{c}}_{i}}}}(t)=1$ task. If the transmission rate ${{r}_{{{{c}}_{i}}}}(t)$ within a time slot is greater than the task size, i.e., ${{r}_{{{{c}}_{i}}}}(t)\cdot \tau \ge {{d}_{{{{c}}_{i}},1}}$, then ${{\phi }_{{{{c}}_{i}},1}}(t)$ increases by the transmission time $\frac{{{d}_{{{{c}}_{i}},1}}\left( t \right)}{{{r}_{{{{c}}_{i}}}}(t)}$. Otherwise the ${{J}_{{{{c}}_{i}}}}(t)=1$ task only wait for the next time slot to be sent and ${{\phi }_{{{{c}}_{i}},1}}(t)$ increases by $\tau $. For other tasks, i.e., ${{J}_{{{{c}}_{i}}}}(t)>1$ tasks, their AoI ${{\phi }_{{{{c}}_{i}},{{J}_{{{{c}}_{i}}}}}}(t)$ are calculated as ${{\phi }_{{{{c}}_{i}},{{J}_{{{{c}}_{i}}}}}}(t)={{\phi }_{{{{c}}_{i}},{{J}_{{{{c}}_{i}}}}}}\left( t-1 \right)+\tau $.

The average AoI of each vehicle can be obtained by calculating the average AoI of all tasks in the queue, i.e,
\begin{equation}
\overline{{{\phi }_{{{{c}}_{i}}}}}(t)=\frac{1}{{{M}_{{{{c}}_{i}}}}(t)}\sum\limits_{j=0}^{{{M}_{{{{c}}_{i}}}}(t)}{{{\phi }_{{{{c}}_{i}},{{J}_{{{{c}}_{i}}}}}}}(t),
\label{eq8}
\end{equation}
where ${{M}_{{{{c}}_{i}}}}(t)$ is the number of tasks of vehicle ${{{c}}_{i}}$ at time slot $t$. 
The system's average AoI is caculated as 
\begin{equation}
\overline{\phi }(t)=\frac{1}{{{N}_{c}}(t)}\sum\limits_{i=1}^{{{N}_{c}}(t)}{\overline{{{\phi }_{{{{c}}_{i}}}}}(t)},
\label{eq9}
\end{equation}
where ${{N}_{c}}(t)$ is the number of vehicles within the RSU coverage area at time slot $t$. Our research problem is to optimize the system AoI within the coverage area of RSUs. 
\section{Cooperative Task Offloading Scheme}
\label{scheme}
In this section, we introduce the Federated Graph Neural Network Multi-Agent Reinforcement Learning (FGNN-MADRL) algorithm. We start by presenting a FL algorithm based on GNNs. Then we formulate the MADRL framework. Finally, we employ the GNN combined Multi-agent SAC algorithm for cooperative task offloading among vehicles.

\begin{figure*}[ht]
	\center
	\includegraphics[scale=0.4]{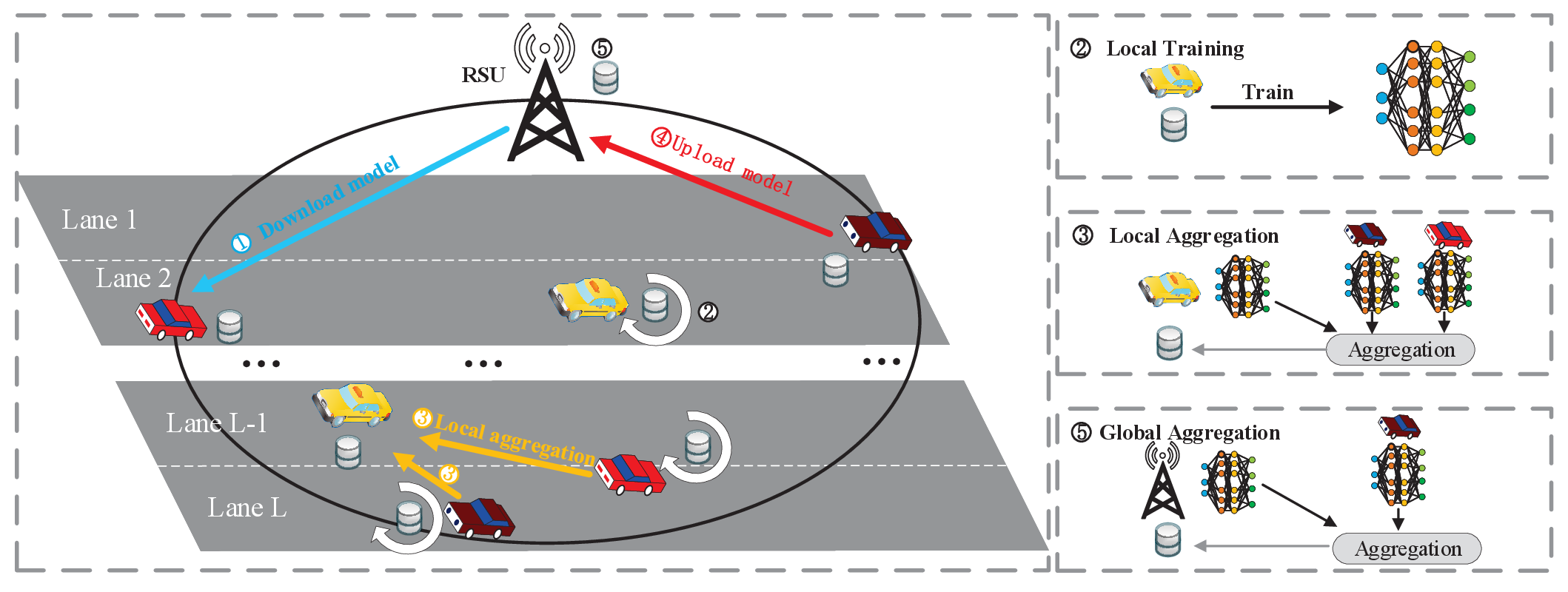}
	\caption{Distributed FL Based on GNN.}
	\label{fig_fl}
\end{figure*}
\subsection{FL Algorithm Based on GNN}
FL facilitates collaborative training of DRL models, while each vehicle retains its training data. Through federated aggregation, vehicles can share knowledge and experience with each other, enhancing the performance and effectiveness of the global model. This decentralized approach significantly alleviates privacy concerns and reduces communication overhead associated with centralized training methods. 
As shown in Fig. \ref{fig_fl}, the distributed FL algorithm based on GNN performs ${{R}^{\max }}$ rounds, each consisting of the following four steps:

\subsubsection {Download Model} In the model, once vehicles enter the RSU coverage area, they download the latest DRL model from the RSU. The DRL model adopted  is based on the Actor-Critic framework. Therefore, at time slot $t$, vehicles download the latest global actor network model $\omega _{a}^{global}(t)$, global critic network model $\omega _{c}^{global}(t)$, and global target critic network model $\omega _{tc}^{global}(t)$ as their local initial models $\omega _{a}^{{{{c}}_{i}}}(t)$, $\omega _{c}^{{{{c}}_{i}}}(t)$, and $\omega _{tc}^{{{{c}}_{i}}}(t)$.

\subsubsection {Local Training} Vehicle ${{{c}}_{i}}$ begins the local training after downloading the global DRL models from the RSU. 
This process involves interacting with the environment to collect training data and storing them in a replay buffer ${{\mathcal{B}}_{{{c}_{i}}}}$ with a certain capacity ${{D}_{s}}$. The local model undergoes ${{I}_{{{c}_{i}}}}$ iterations of updating, each can be represented as
\begin{equation}
\omega _{r}^{{{c}_{i}}}\leftarrow \omega _{r}^{{{c}_{i}}}-\eta _{r}^{{{c}_{i}}}\nabla F_{r}^{{{c}_{i}}}\left( \omega _{r}^{{{c}_{i}}} \right),r\in \{a,c,tc\},
\label{eq14}
\end{equation}
where $\nabla F_{r}^{{{c}_{i}}}\left( \omega _{r}^{{{c}_{i}}} \right)$ denotes the gradient of $\omega _{r}^{{{c}_{i}}},r\in \{a,c,tc\}$ and $\eta _{r}^{{{c}_{i}}}$ is a fixed learning rate. Noted that due to differences in vehicular computational resources and the number of training iterations ${{I}_{{{c}_{i}}}}$, each vehicle has a different training time ${{t}_{{{{c}}_{1}}}}$.

\subsubsection {Local Aggregation} Traditional FL involves uploading models to the RSU for global aggregation upon completion of local training. However, frequent model uploads can incur substantial communication overhead, consuming excessive bandwidth and impacting task offloading and AoI. 
 In addition directly employing an average FL for updating local models might overlook the unique characteristics of each vehicle's previously trained model. This not only wastes computational resources consumed during training but also harms model personalization. Therefore, we first perform local model aggregation based on GNNs among vehicles, where GNNs are used to generate the aggregation weights by capturing the characteristics of vehicular network.

\begin{figure*}[ht!]
	\center
	\includegraphics[scale=0.45]{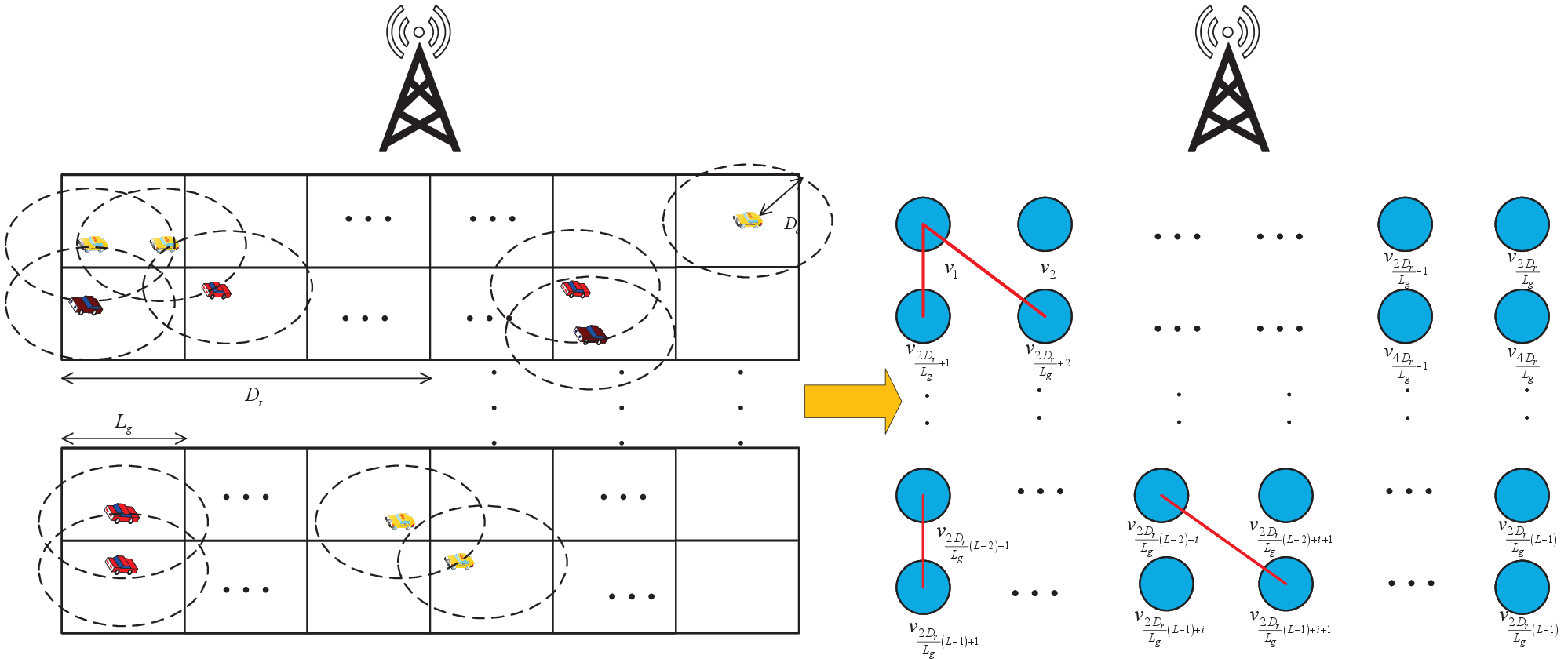}
	\caption{GNN road graph.}
	\label{fig_gnn}
\end{figure*}

Due to the varying number of vehicles within the RSU coverage area, treating each vehicle as a node would increase the complexity of the GNN network. To overcome this issue, we divide the road into multiple segments, with each segment acting as a node of the GNN. As illustrated in Fig. \ref{fig_gnn}, We define the road within the coverage area of a single RSU as a graph $G(t)=(V(t),E(t))$, where each node $V(t)$ represents a road segment with length ${{L}_{g}}$ and the total number of nodes is $\frac{2{{D}_{r}}}{{{L}_{g}}}L$. 
For any given node ${{v}_{i}}\in V(t)$, its feature  ${{\psi }_{{{v}_{i}}}}\left( t \right)$ is defined as
\begin{equation}
{{\psi }_{{{v}_{i}}}}\left( t \right)=[{{n}_{{{v}_{i}}}}(t),{{a}_{{{v}_{i}}}}(t),{{\mathcal{L}}_{{{v}_{i}},a}}(t),{{\mathcal{L}}_{{{v}_{i}},c}}(t),{{\mathcal{L}}_{{{v}_{i}},tc}}(t)]
\label{eq15},
\end{equation}
where ${{n}_{{{v}_{i}}}}(t)$ represents the number of vehicles in node ${{v}_{i}}$ at time slot $t$, and ${{a}_{{{v}_{i}}}}(t)$ represents the average number of times all vehicles within node ${{v}_{i}}$ have participated in local aggregation. Furthermore, ${{\mathcal{L}}_{{{v}_{i}},a}}(t)$, ${{\mathcal{L}}_{{{v}_{i}},c}}(t)$ and ${{\mathcal{L}}_{{{v}_{i}},{{tc}}}}(t)$ represents the average loss values of the actor network, critic network and critic network of all vehicles within node ${{v}_{i}}$, respectively.

We define a set ${{\Omega }_{{{v}_{i}}}}\left( t \right)=\bigcup\limits_{{{c}_{j}}\in {{\mathcal{C}}_{{{c}_{i}}}}\left( t \right)}{N}({{c}_{j}})$ containing all nodes within the communication range ${{{D}}_{c}}$ of vehicles in node ${{v}_{i}}$, where ${{\mathcal{C}}_{{{c}_{i}}}}\left( {t} \right)$ denotes the set of vehicles in node ${{v}_{i}}$, and $N({{c_j}})$ denotes the set of nodes formed by the vehicles within the communication range of vehicle ${{c_j}}$. 
For each node ${{v}_{j}}$ in ${{\Omega }_{{{v}_{i}}}}\left( t \right)$, we establish an edge between nodes ${{v}_{i}}$ and ${{v}_{j}}$ and the edge set $E(t)$ of the graph $G(t)$ can be obtained. 
For example, if node ${{v}_{1}}$ contains two vehicles, both with a communication range of ${{{D}}_{c}}$, and the vehicles within the communication range of vehicle ${{c}_{1}}$ are in node ${{v}_{\frac{2{{D}_{r}}}{{{L}_{g}}}+1}}$, while the vehicles within the communication range of vehicle ${{c}_{2}}$ are in node ${{v}_{\frac{2{{D}_{r}}}{{{L}_{g}}}+2}}$, then node ${{v}_{1}}$ forms undirected edges with nodes ${{v}_{\frac{2{{D}_{r}}}{{{L}_{g}}}+1}}$ and ${{v}_{\frac{2{{D}_{r}}}{{{L}_{g}}}+2}}$. Similarly, edges for other nodes are constructed in the same way. We then define the adjacency matrix $A(t)$ of graph $G(t)$ with dimensions $\frac{2{{D}_{r}}}{{{L}_{g}}}L\times \frac{2{{D}_{r}}}{{{L}_{g}}}L$. If $({{v}_{i}},{{v}_{j}})\in E(t)$, then ${{A}_{ij}(t)}=1$, otherwise ${{A}_{ij}(t)}=0$.

Next, we transform the feature vector ${{\psi }_{{{v}_{i}}}}\left( t \right)$ of node ${{v}_{i}}\in V(t)$. The intermediate representation of node ${{v}_{i}}$ in layer ${{l}_{gnn}}$ of
 GNN is denoted as $\bar{h}_{i}^{{{l}_{gnn}}}$ and is calculated as
\begin{equation}
\bar{h}_{i}^{{{l}_{gnn}}}={{\mathbf{W}}_{g}}^{{{l}_{gnn}}}\cdot h_{i}^{{{l}_{gnn}}-1}
\label{eq16},
\end{equation}
where ${{\mathbf{W}}_{g}}^{{{l}_{gnn}}}$ is the learnable transformation matrix for layer ${{l}_{gnn}}$, and $h_{i}^{{{l}_{gnn}}-1}$ is the representation of node ${{v}_{i}}$ from the previous layer, with the first layer representation $h_{i}^{0}$ being the input feature vector ${{\psi }_{{{v}_{i}}}}\left( t \right)$.
We then aggregate neighborhood information for $\bar{h}_{i}^{{{l}_{gnn}}}$ as 
\begin{equation}
h_{i}^{{{l}_{gnn}}}=\sigma \left( \bar{h}_{i}^{{{l}_{gnn}}}+\sum\limits_{j\in {{\Omega }_{{{v}_{i}}}}\left( t \right)}{{{\varepsilon }_{ij}}} \bar{h}_{j}^{{{l}_{gnn}}} \right),
\label{eq17}
\end{equation}
where $\bar{h}_{i}^{{{l}_{gnn}}}+\sum\limits_{j\in {{\Omega }_{{{v}_{i}}}}\left( t \right)}{{{\varepsilon }_{ij}}} \bar{h}_{j}^{{{l}_{gnn}}}$ represents the intermediate representation of other aggregated nodes, ${{\varepsilon }_{ij}}$ is the aggregation weight, and $\sigma (\cdot )$ is a nonlinear activation function. 

Each GNN network layer operates as above, ultimately outputting an encoded graph ${{G}^{(H)}}(t)\in {{\mathbb{R}}^{\frac{2{{R}_{r}}}{{{L}_{g}}}L\times p}}$ that includes feature embedding vectors for all nodes ${{\bar{\psi }}_{{{v}_{i}}}}\left( t \right)\in {{\mathbb{R}}^{p}}$, where $p$ is the dimension of the feature space, and $H$ denotes the number of layers in the GNN. We represent the set of all node-extracted feature vectors as $\bar{\psi }\left( t \right)=[{{\bar{\psi }}_{{{v}_{1}}}}\left( t \right),{{\bar{\psi }}_{{{v}_{2}}}}\left( t \right)...{{\bar{\psi }}_{{{v}_{i}}}}\left( t \right)]$.

 
 We define a set ${{B}_{{{{c}}_{i}}}}\left( t \right)=\left\{ {{{\bar{\psi }}}_{{{v}_{j}}}}(t)|{{v}_{j}}\in N({{{c}}_{i}}) \right\}$ containing the extracted node features of all vehicles within the communication range of vehicle $c_i$, where $N({{{c}}_{i}})$ represents the set of nodes containing vehicles within the communication range of vehicle ${{{c}}_{i}}$. We then determine the weights for model aggregation by performing a softmax on all extracted node features in ${{B}_{{{{c}}_{i}}}}\left( t \right)$, i.e.,
\begin{equation}
{{\alpha }_{{{c}_{i}}}}\left( t \right)=\frac{{{e}^{{{{\bar{\psi }}}_{{{v}_{i}}}}(t)}}}{{{e}^{{{{\bar{\psi }}}_{{{v}_{i}}}}(t)}}+\sum\limits_{{{v}_{j}}\in {{B}_{{{c}_{i}}}}(t)}{{{e}^{{{{\bar{\psi }}}_{{{v}_{j}}}}(t)}}}},
\label{eq18}
\end{equation}
where ${{\alpha }_{{{c}_{i}}}}\left( t \right)$ represents the aggregation weight of vehicle ${{{c}}_{i}}$.

Inspired by the centralized critic network concept of Multi-Agent Deep Deterministic Policy Gradient (MADDPG), our model aggregates only the critic network during local aggregation. This design allows the critic to evaluate the value of its own actions in consideration of other vehicles' behaviors, as the value often depends on the overall environment. Besides, each vehicle's actor network can independently make decisions based on its experience and environment. Assuming the latest critic model parameter for vehicle ${{{c}}_{i}}$ at time slot $t$ is $\omega _{c}^{{{{c}}_{i}}}(t)$. Hence, the critic model aggregation for vehicle ${{{c}}_{i}}$ is caculated as
\begin{equation}
\omega _{c}^{{{{c}}_{i}}}(t)={{\alpha }_{{{c}_{i}}}}\left( t \right) \omega _{c}^{{{{c}}_{i}}}(t)+\sum\limits_{{{c}_{j}}\in {{\mathcal{C}}_{{{c}_{i}}}}\left( {t} \right)}{{{\alpha }_{{{c}_{j}}}}\left( t \right) \omega _{c}^{{{c}_{j}}}(t)}
\label{eq19},
\end{equation}
where ${{\mathcal{C}}_{{{c}_{i}}}}\left( {t} \right)$ represents the set of vehicles within the communication range of vehicle ${{{c}}_{i}}$. After completing local aggregation, vehicles start the next round of local training.
\subsubsection {Upload Model} Before leaving the RSU coverage area, vehicle ${{{c}}_{i}}$ uploads its latest trained actor network model $\omega _{a}^{{{{c}}_{i}}}(t)$ critic network model $\omega _{c}^{{{{c}}_{i}}}(t)$, and target critic network model $\omega _{tc}^{{{{c}}_{i}}}(t)$ to the RSU.
\subsubsection {Global Aggregation} After receiving the local models from all vehicles about to exit its coverage range, the RSU performs asynchronous federated aggregation. We use $\overline{\mathcal{C}}\left( {t} \right)$ to denote the set of all vehicles leaving the RSU coverage area at time slot $t$. The RSU computes the new global actor network model $\omega _{a}^{global}(t)$, global critic network model $\omega _{c}^{global}(t)$ and global target critic network model $\omega _{tc}^{global}(t)$ by
\begin{equation}
\omega _{r}^{global}\leftarrow \sum\limits_{\overline{{{c}_{i}}}\in \overline{\mathcal{C}}\left( {t} \right)}{\frac{\omega _{r}^{\overline{{{c}_{i}}}}(t)}{\left| \overline{\mathcal{C}}\left( {t} \right) \right|}},r\in \{a,c,tc\}
\label{eq20},
\end{equation}
where $\omega _{r}^{\overline{{{c}_{i}}}}(t),r\in \{a,c,tc\}$ represents the latest model of disappearing vehicle $\overline{{{c}_{i}}}$ at time slot $t$. Then the federated aggregation process is completed, and the RSU has obtained an updated global model. 
However, model uploading requires additional bandwidth resources. We assume the size of the model as $|{{W}_{{{{c}}_{i}}}}|$ and denote the minimum transmission power required for vehicle ${{{c}}_{i}}$ to upload its model in the current time slot as ${{p}_{{{{c}}_{i}},\omega }}(t)$, with the number of vehicles that need to upload their models in the current time slot being ${{N}_{m}}(t)$. Therefore, the transmission rate for offloading tasks of each vehicle is recalculated as
\begin{equation}
\begin{aligned}
&{{R}_{{{{c}}_{i}}}}(t)= \\
&B {{\log }_{2}}\left( 1+\frac{{{p}_{{{{c}}_{i}}}}(t){{g}_{{{{c}}_{i}}}}(t)}{\sum\limits_{j\ne i}^{N(t)}{{{p}_{{{{c}}_{j}}}}}(t) {{g}_{{{{c}}_{j}}}}+\sum\limits_{k=1}^{{{N}_{m}}(t)}{{{p}_{{{{c}}_{k}},\omega }}}(t) {{g}_{{{{c}}_{k}}}}(t)+{{\sigma }^{2}}} \right),
\end{aligned}
\label{eq10}.
\end{equation}
Thus ${{\phi }_{{{{c}}_{i}},1}}(t)$ of the ${{J}_{{{{c}}_{i}}}}(t)=1$ task is also recalculated as
\begin{equation}
{{\phi }_{{{{c}}_{i}},1}}\left( t \right)=\left\{ \begin{matrix}
   {{\phi }_{{{{c}}_{i}},1}}\left( t-1 \right)+\frac{{{d}_{{{{c}}_{i}},1}}\left( t \right)}{{{R}_{{{{c}}_{i}}}}(t)},{ if }{{{R}}_{{{{c}}_{i}}}}(t)\cdot \tau \ge {{d}_{{{{c}}_{i}},1}}{                   }  \\
   {{\phi }_{{{{c}}_{i}},1}}\left( t-1 \right)+\tau ,{ otherwise                             }  \\
\end{matrix} \right.
\label{eq11}.
\end{equation}
The process of the FL algorithm is shown in Algorithm \ref{algorithm1}. Subsequently, the RSU sends the updated global model to all new coming vehicles entering the coverage area.

\begin{algorithm}[b!]
 	\caption{Distributed FL Algorithm Based on GNN}
 	\label{algorithm1}
	Initialize the global models: $\omega _{a}^{global}(0)$, $\omega _{c}^{global}(0)$, and $\omega _{tc}^{{{{c}}_{i}}}(0)$\;
	
 	Initialize the set of vehicles: \({{\mathbb{V}}_{t}}=\{\}\)\;
 	
 	\For{t from 1 to \({{R}^{\max }}\)}
 	{
 		
 		\For{each vehicle \({{c}_{i}}\) in \({{\mathbb{V}}_{t}}\) in parallel}
 		{
 			Update the position: ${{x}_{{{c}_{i}}}}(t)\leftarrow {{x}_{{{c}_{i}}}}(t)+v_{\max }^{{{l}_{idx}}}$\;
 			 \If{vehicle \({{c}_{i}}\) exits  the RSU coverage area}
 			{
 				Upload models: $\omega _{a}^{{{{c}}_{i}}}(t)$, $\omega _{c}^{{{{c}}_{i}}}(t)$, $\omega _{tc}^{{{{c}}_{i}}}(t)$\;
 				Update the global model based on Eq. \eqref{eq20}\;
 				Remove vehicle \({{c}_{i}}\) from the set ${{\mathbb{V}}_{t}}$: ${{\mathbb{V}}_{t}}\leftarrow {{\mathbb{V}}_{t}}\backslash \{{{c}_{i}}\}$\;
 			}
 		}
 		\If{vehicle \({{c}_{i}}\) enters the RSU coverage area}
 		{
 				Download models: $\omega _{a}^{{{{c}}_{i}}}(t)$, $\omega _{c}^{{{{c}}_{i}}}(t)$, $\omega _{tc}^{{{{c}}_{i}}}(t)$ \(\leftarrow\) $\omega _{a}^{global}(t)$, $\omega _{c}^{global}(t)$, $\omega _{tc}^{{{{c}}_{i}}}(t)$\;
 				Add vehicle \({{c}_{i}}\) to the set ${{\mathbb{V}}_{t}}$: ${{\mathbb{V}}_{t}}\leftarrow {{\mathbb{V}}_{t}}\bigcup \{{{c}_{i}}\}$\;
 		}
 		Update the vehicle road graph structure $G(t)=(V(t),E(t))$\;
 		\For{each vehicle \({{c}_{i}}\) in \({{\mathbb{V}}_{t}}\) in parallel}
 		{
 			Update the position: ${{x}_{{{c}_{i}}}}(t)\leftarrow {{x}_{{{c}_{i}}}}(t)+v_{\max }^{{{l}_{idx}}}$\;
 			 \If{training for vehicle \({{c}_{i}}\) is completed}
 			{
 				Calculate the local model aggregation weight ${{\alpha }_{{{c}_{i}}}}\left( t \right)$ extracted by GNN based on Eq. \eqref{eq19}\;
 				Update the local model based on Eq. \eqref{eq20}\;
 			}
 		}
 		Train the GNN network\;		
 	}
 Return $\omega _{a}^{global}(t)$, $\omega _{c}^{global}(t)$, $\omega _{tc}^{{{{c}}_{i}}}(t)$.
\end{algorithm}
\subsection{Cooperative offloading Scheme Based on MADRL}

Next, the DRL framework is first formulated, which is the basis of the MADRL algorithm. Then, the GNN combined MASAC algorithm will be introduced.
\subsubsection{MADRL Framework} The MADRL framework includes states, actions and rewards which are defined as follows.

\paragraph{States}
In our model, the state of each vehicle comprises several key factors,i.e.,
\begin{equation}
{{s}_{{{c}_{i}}}}(t)=[{{g}_{{{c}_{i}}}}\left( t \right),{{\phi }_{{{{c}}_{i}},0}}\left( t \right),\overline{\phi }(t),{{l}_{{{c}_{i}},r}}(t),{{d}_{{{{c}}_{i}},0}},{{N}_{c}}(t)]
\label{eq21},
\end{equation}
where ${{g}_{{{c}_{i}}}}\left( t \right)$ is the channel gain, an important indicator of communication quality between the vehicle and RSU. ${{\phi }_{{{{c}}_{i}},0}}\left( t \right)$ is the AoI of the task waiting to be sent. ${{l}_{{{c}_{i}},r}}(t)$ indicates the distance of vehicle ${{{c}}_{i}}$ from the RSU. ${{d}_{{{{c}}_{i}},0}}$ is the size of the task waiting to be sent and ${{N}_{c}}(t)$ represents the number of vehicles within the RSU coverage. The RSU monitors the environment, including the system's average AoI $\overline{\phi }(t)$ and the number of vehicles ${{N}_{c}}(t)$, and sends these information to all vehicles through a downlink. We assume that the data sizes of these information are very small, so the delay in transmitting this information can be considered negligible.

\paragraph{Action}
Each vehicle needs to decide its transmission power for transmitting tasks, i.e.,
\begin{equation}
{{p}_{{{c}_{i}}}}(t)\in [0,{{p}_{{max}}}]
\label{eq22},
\end{equation}
where ${{p}_{{max}}}$ is the maximum transmission power for each vehicle. The transmission power of vehicles interferes with and affects the transmission rate of other vehicles.

\paragraph{Reward}
The purpose of the reward is to optimize the behavior of vehicles to minimize the system's average AoI. The reward ${{r}_{{{c}_{i}}}}(t)$ for vehicle ${{{c}}_{i}}$ at time slot $t$ is designed as

\begin{equation}
{{r}_{{{c}_{i}}}}(t)=\left\{ \begin{array}{*{35}{l}}
   -\left( \bar{\phi }(t)+{{p}_{{{c}_{i}}}}(t)\cdot {{\omega }_{0}}+\xi (t)\cdot {{\omega }_{{2}}} \right), & {{M}_{{{c}_{i}}}}(t)>0  \\
   -\left( \bar{\phi }(t)+{{p}_{{{c}_{i}}}}(t)\cdot {{\omega }_{1}}+\xi (t)\cdot {{\omega }_{{2}}} \right), & {{M}_{{{c}_{i}}}}(t)=0  \\
\end{array} \right.
\label{eq23}.
\end{equation}
where ${{M}_{{{{c}}_{i}}}}(t)$ is the number of tasks for vehicle ${{{c}}_{i}}$ at time slot $t$. ${{\omega }_{0}}=1+\frac{\bar{\phi }(t)}{{{\phi }_{{{{c}}_{i}},0}}\left( t \right)}$ and ${{\omega }_{1}}=1+\bar{\phi }(t)$ are factors representing the impact of the vehicle's own power on the system's average AoI. The inclusion of ${{p}_{{{c}_{i}}}}(t)\cdot {{\omega }_{0}}$ and ${{p}_{{{c}_{i}}}}(t)\cdot {{\omega }_{1}}$ in the reward considers the interference caused by the vehicle's transmission power to others. Higher transmission power ${{p}_{{{c}_{i}}}}(t)$ causes significant interference to others, resulting in a larger penalty in the reward function. Conversely, a relatively high AoI with lower transmission power ${{p}_{{{c}_{i}}}}(t)$ also leads to a substantial penalty. The addition of 1 in ${{\omega }_{0}}$ and ${{\omega }_{1}}$ ensures that ${{p}_{{{c}_{i}}}}(t)\cdot {{\omega }_{0}}$ and ${{p}_{{{c}_{i}}}}(t)\cdot {{\omega }_{1}}$ do not equal zero when $\frac{\bar{\phi }(t)}{{{\phi }_{{{{c}}_{i}},0}}\left( t \right)} = 0$ and $\bar{\phi }(t) = 0$, avoiding abrupt value changes that may likely cause instability in model training.

${{\omega }_{{2}}}$ is a weight factor and a hyperparameter. $\xi (t)$ represents the penalty for unprocessed tasks when the vehicle leaves the RSU coverage area. It is designed to encourage vehicles to offload as many tasks as possible within the RSU.
Otherwise, the trained model will rely on the disappearance of vehicles to reduce the Average AoI, resulting in vehicles maintaining a transmission power of zero within the RSU coverage area. $\xi (t)$ is a recursive function which is calculated as
\begin{equation}
\xi (t)=\left\{ \begin{array}{*{35}{l}}
   \xi (t-1)+\frac{1}{{{N}_{{c}}}\left( t \right)}\sum\limits_{\overline{{{c}_{i}}}\in \overline{\mathcal{C}}\left( {t} \right)}{\overline{{{\phi }_{\overline{{{c}_{i}}}}}}}(t) & ,\left| \overline{\mathcal{C}}\left( {t} \right) \right|>0  \\
   \xi (t-1)\times \delta  & ,\left| \overline{\mathcal{C}}\left( {t} \right) \right|=0  \\
\end{array} \right.
\label{eq24},
\end{equation}
where $\overline{\mathcal{C}}\left( {t} \right)$ denotes the set of all vehicles about to leave the RSU coverage area at time slot $t$, $\left| \overline{\mathcal{C}}\left( {t} \right) \right|$ is the number of vehicles in the set, $\overline{{{\phi }_{\overline{{{c}_{i}}}}}}\left( t \right)$ is the AoI of the disappearing vehicle $\overline{{{c}_{i}}}$, and $\delta \in (0,1]$ is the decay factor for the penalty term, indicating a continuous and gradually diminishing impact of the penalty after vehicles leave the RSU range.
The expected long-term discounted reward for vehicle ${{{c}}_{i}}$ is calculated as
\begin{equation}
J\left( {{\mu }_{{{c}_{i}}}} \right):={{\mathbb{E}}_{{{\mu }_{{{c}_{i}}}}}}\left[ \sum\limits_{t=1}^{{{N}_{c}}(t)}{{{\gamma }^{t-1}}}{{r}_{{{c}_{i}}}}(t) \right]
\label{eq25},
\end{equation}
where $\gamma \in [0,1]$ is the discount factor, and ${{N}_{c}}(t)$ represents the number of vehicles within the RSU coverage at time slot $t$. 
Our goal is to find the optimal strategy $\mu _{{{c}_{i}}}^{*}$ that maximizes the expected long-term discounted reward for vehicle ${{{c}}_{i}}$. 

\subsubsection{GNN combined  MASAC algorithm}
In the considered scenario, due to the continuous action space of vehicle transmission power, the RL model for each vehicle employs the SAC model. Compared to DDPG, SAC shows better sample efficiency and stability. SAC introduces entropy regularization into the RL framework, encouraging exploration and achieving more robust policy learning. This is particularly beneficial in the dynamic and complex vehicular network environment. According to the SAC algorithm, the expected long-term discounted reward is calculated as
\begin{equation}
\begin{aligned}
&J\left( {{\pi }_{{{c}_{i}},t}}\left( {{a}_{{{c}_{i}},t}}|{{s}_{{{c}_{i}},t}} \right) \right)={{E}_{\tau \sim{{\pi }_{{{c}_{i}},t}}\left( {{a}_{{{c}_{i}},t}}|{{s}_{{{c}_{i}},t}} \right)}} \\
&\left[ \sum\limits_{t=0}^{T}{{{\gamma }^{t-1}}}{{r}_{{{c}_{i}}}}(t)+{{\beta }_{{{c}_{i}},t}}H\left( {{\pi }_{{{c}_{i}},t}}\left( {{a}_{{{c}_{i}},t}}|{{s}_{{{c}_{i}},t}} \right) \right) \right]
\end{aligned}
\label{eq26},
\end{equation}
where ${{s}_{{{c}_{i}},t}}$, ${{a}_{{{c}_{i}},t}}$, and ${{\pi }_{{{c}_{i}},t}}\left( {{a}_{{{c}_{i}},t}}|{{s}_{{{c}_{i}},t}} \right)$ respectively represent the state, action and strategy for vehicle ${{c}_{i}}$ at time slot $t$. $H\left( {{\pi }_{{{c}_{i}},t}}\left( {{a}_{{{c}_{i}},t}}|{{s}_{{{c}_{i}},t}} \right) \right)$ is the entropy of the policy. ${{\beta }_{{{c}_{i}},t}}$ is a balancing factor between exploring feasible strategies and maximizing rewards for vehicle ${{c}_{i}}$, which is dynamically adjusted based on the state ${{s}_{{{c}_{i}},t}}$. The optimal $\beta _{_{{{c}_{i}},t}}^{*}$ under state ${{s}_{{{c}_{i}},t}}$ is defined as
\begin{equation}
\beta _{_{{{c}_{i}},t}}^{*}=\arg {{\min }_{{{\beta }_{{{c}_{i}},t}}}}{{E}_{{{a}_{t}}\sim\pi _{t}^{*}}}\left[ -{{a}_{{{c}_{i}},t}}\log \pi _{{{c}_{i}},t}^{*}-{{\beta }_{{{c}_{i}},t}}{{\overline{H}}_{{{c}_{i}},t}} \right]
\label{eq27},
\end{equation}
where ${{\overline{H}}_{{{c}_{i}},t}}=dim({{a}_{{{c}_{i}},t}})$ represents the dimension of the action. $\pi _{{{c}_{i}},t}^{*}$ is the optimal strategy for vehicle ${{c}_{i}}$ at time slot $t$, which is calculated as
\begin{equation}
\pi _{{{c}_{i}},t}^{*}=\arg \underset{{{\pi }_{{{c}_{i}},t}}}{\mathop{\max }}\,J\left( {{\pi }_{{{c}_{i}},t}}\left( {{a}_{{{c}_{i}},t}}|{{s}_{{{c}_{i}},t}} \right) \right)
\label{eq28},
\end{equation}
The SAC algorithm architecture includes an actor network, two critic networks and two target critic networks. The actor network is responsible for policy improvement, while the two critic networks perform policy evaluation. Two target critic networks aim to improve training speed and stability. Through continuous policy improvement and evaluation, the policy $\pi (t)$ eventually converges to the optimal policy ${{\pi }^{*}}(t)$.

To evaluate the quality of weights generated by GNNs, we leverage the policy gradient method. Specifically, the RSU maintains a GNN network and a GNN critic network. The GNN critic network evaluates the performance of generated weights, while the GNN network improves its weight generation capability based on feedback from the GNN critic network. Moreover, for more stable target values, we introduce a target GNN critic network. Next the detailed process of the FGNN-MADRL algorithm will be  explained.

\paragraph{Training Stage}
Let ${{\theta }_{a}}$ represent the parameters of the actor network, ${{\varphi }_{1}}$ and ${{\varphi }_{2}}$ represent the parameters of the two critic networks, and ${{\bar{\varphi }}_{1}}$ and ${{\bar{\varphi }}_{2}}$ represent the parameters of the two Target critic networks. The pseudocode for the SAC training phase algorithm is shown in Algorithm \ref{algorithm2}.

Initially, RSU randomly initialize the SAC model parameters, including two global critic network parameters $\varphi _{1}^{global}$ and $\varphi _{2}^{global}$, global actor network parameter $\theta _{a}^{global}$, two global target critic network parameters $\bar{\varphi }_{1}^{global}$ and $\bar{\varphi }_{2}^{global}$ (initialized same to $\varphi _{1}^{global}$ and $\varphi _{2}^{global}$), and ${{\beta }^{{g}lobal}}$. We also initialize the GNN network parameter ${{\theta }_{g}}$, critic GNN critic parameter ${{\varphi }_{g}}$, and GNN target critic model parameters ${{\bar{\varphi }}_{g}}$. A GNN experience replay buffer ${{\mathcal{B}}_{g}}$ with a storage capacity ${{D}_{g}}$ is set up in the RSU to store the road's graph data.

First  all vehicles  are cleared on the road. Then a new road graph $G(0)$ is generated based on the current road scenario, where each node's feature vector is a zero vector and there are no edges between nodes. We input $G(0)$ into the GNN network to obtain an initial  extracted feature vectors $\bar{\psi }\left( 0 \right)$. Next, we recalculate the average AoI $\bar{\phi }(0)$ within the RSU coverage area. To simulate the dynamic movement of vehicles, we generate the first vehicle entry into the RSU coverage area at time slots according to a Poisson distribution ${{\lambda }_{L}}$.

The algorithm simulates from time slot 1 to $R^{\max}$. In each time slot, vehicles update their positions based on their speed and perform boundary checks to determine if any vehicles have left the RSU coverage area. Vehicles leaving the RSU coverage upload their latest local model parameters $\varphi _{1}^{{{c}_{i}}}(t)$, $\varphi _{2}^{{{c}_{i}}}(t)$, $\theta _{a}^{{{c}_{i}}}(t)$, $\bar{\varphi }_{1}^{{{c}_{i}}}(t)$, $\bar{\varphi }_{2}^{{{c}_{i}}}(t)$ to the RSU for global federated averaging. Additionally, each vehicle calculates the minimum power ${{p}_{{{{c}}_{i}},\omega }}(t)$ which is required to upload the model. New vehicles entering the RSU coverage area are added to the system and download the latest model from the RSU. They also initialize their SAC experience replay buffer ${{\mathcal{B}}_{{{c}_{i}}}}$ with a certain storage capacity ${{D}_{s}}$.

Each vehicle checks for new tasks in the current time. If a new task is generated, a task size uniformly distributed within $[{{d}_{\min }},{{d}_{\max }}]$ is produced and stored in the vehicle's task queue, with the next task arrival interval generated according to a Poisson distribution $\mu $. Vehicles observe the current state ${{s}_{{{c}_{i}}}}(t)$, i.e., Eq. \eqref{eq21}, then inputs ${{s}_{{{c}_{i}}}}(t)$ into its actor network to generate its action ${{p}_{{{c}_{i}}}}(t)$. Based on each vehicle's channel gain ${{g}_{{{c}_{i}}}}\left( t \right)$, transmission power ${{p}_{{{c}_{i}}}}(t)$ and model transmission power ${{p}_{{{{c}}_{i}},\omega }}(t)$, the transmission rate ${{R}_{{{{c}}_{i}}}}(t)$ is calculated based on Eq. \eqref{eq10}. Each vehicle then executes task offloading. The RSU computes the next time slot's average AoI $\bar{\phi }(t+1)$ and each vehicle computes their own average AoI ${{\phi }_{{{{c}}_{i}},0}}\left( t+1 \right)$ based on Eq. \eqref{eq8} and Eq. \eqref{eq9}. Each vehicle's reward ${{r}_{{{c}_{i}}}}(t)$ is calculated based on Eq. \eqref{eq24}.

\begin{algorithm}[b!]
 	\caption{GNN Combined SAC Training Stage Algorithm}
 	\label{algorithm2}
 	\KwIn{$\theta _{a}^{global}$, $\varphi _{1}^{global}$, $\varphi _{2}^{global}$, ${{\beta }^{{global}}}$, ${{\beta }^{{global}}}$}
 	\KwOut{optimized ${{(\theta _{a}^{global})}^{*}}$}
	Randomly initialize models: $\theta _{a}^{global}$, $\varphi _{1}^{global}$, $\varphi _{2}^{global}$, ${{\beta }^{{global}}}$, ${{\beta }^{{global}}}$\;
	Initialize models: $\bar{\varphi }_{1}^{global}\leftarrow \varphi _{1}^{global}$, $\bar{\varphi }_{2}^{global}\leftarrow \varphi _{2}^{global}$\;
	Initialize the set of vehicles: \({{\mathbb{V}}_{t}}=\{\}\)\;
	\For{each vehicle \({{c}_{i}}\) in \({{\mathbb{V}}_{t}}\) in parallel}
	{
		Update positions in vehicle set \({{\mathbb{V}}_{t}}\)\;
		Update \({{\mathbb{V}}_{t}}\) based on vehicles entering and exiting the RSU coverage area\;
		Generate a new graph data structure $G(t)$\;
		Generate feature vector set $\bar{\psi }\left( t \right)$ based on \eqref{eq15}\;
		\For{each vehicle \({{c}_{i}}\) in \({{\mathbb{V}}_{t}}\) in parallel}
 		{
 			Observe state ${{s}_{{{c}_{i}}}}(t)$ and choose action ${{p}_{{{c}_{i}}}}(t)$\;
 		}
 		\For{each vehicle \({{c}_{i}}\) in \({{\mathbb{V}}_{t}}\) in parallel}
 		{
 			Calculate ${{R}_{{{{c}}_{i}}}}(t)$ based on Eq. \eqref{eq10}\;
 			Calculate $\overline{{{\phi }_{{{{c}}_{i}}}}}(t)$ based on Eq. \eqref{eq8}\;
 			
 		}
 		Calculate $\overline{\phi }(t)$ of the system based on Eq. \eqref{eq9}\;		
 		\For{each vehicle \({{c}_{i}}\) in \({{\mathbb{V}}_{t}}\) in parallel}
 		{
 			Calculate reward ${{r}_{{{c}_{i}}}}(t)$ based on Eq. \eqref{eq23}\;
 			Store $\left( {{s}_{{{c}_{i}}}}(t),{{p}_{{{c}_{i}}}}(t),{{r}_{{{c}_{i}}}}(t),{{s}_{{{c}_{i}}}}(t+1) \right)$ in  ${{\mathcal{B}}_{{{c}_{i}}}}$\;
 			\If{$|{{\mathcal{B}}_{{{c}_{i}}}}|$ \(\ge\) ${{\mathcal{I}}_{{{c}_{i}}}}$}
 			{
 				\For{i from 1 to ${{I}_{{{c}_{i}}}}$}
 				{
 					Randomly sample $\mathcal{M}$ tuples $\left( {{s}_{{{c}_{i}},i}},{{a}_{{{c}_{i}},i}},{{r}_{{{c}_{i}},i}},{{{{s}'}}_{{{c}_{i}},i}} \right)$ as training data from ${{\mathcal{B}}_{{{c}_{i}}}}$\;
 					 Update ${{\beta }_{{{c}_{i}}}}$ based on Eq. \eqref{eq29}\;
 					 Update ${{\theta }_{{{c}_{i}}}}$ based on Eq. \eqref{eq30}\;
 					 Update $\varphi _{1}^{{{c}_{i}}}$ and $\varphi _{2}^{{{c}_{i}}}$ based on Eq. \eqref{eq32};
 				}
 				\If{${{I}_{{{c}_{i}}}}\%{{\tilde{I}}_{{{c}_{i}}}}==0$}
 				{
 				Update $\bar{\varphi }_{1}^{{{c}_{i}}}$ and $\bar{\varphi }_{2}^{{{c}_{i}}}$  based on Eq. \eqref{eq34};
 				}
 				Local federated aggregation\;
 			}
 		}
 		Store  $\left( G(t),\bar{\psi }\left( t \right),\bar{\phi }(t),G(t+1) \right)$ in  ${{\mathcal{B}}_{g}}$\;
 		\If{\(r\%{{T}_{g}}==0\) and $|{{\mathcal{B}}_{g}}|\ge {{\mathcal{I}}_{g}}$}
 		{
 			\For{i from 1 to ${{I}_{{{c}_{i}}}}$}
 			{
 			Randomly sample ${{\mathcal{M}}_{g}}$ tuples $\left( {{G}_{i}},{{{\bar{\psi }}}_{i}},{{{\bar{\phi }}}_{i}},{{{{G}'}}_{i}} \right)$ as training data from ${{\mathcal{B}}_{g}}$\;
 			Update ${{\theta }_{g}}$ based on Eq. \eqref{eq36}\;
 			Update ${{\varphi }_{g}}$ based on Eq. \eqref{eq37}\;

 			}
 			\If{${{I}_{{{c}_{i}}}}\%{{\tilde{I}}_{g}}==0$}
 			{
 				Update ${{\bar{\varphi }}_{g}}$ based on Eq. \eqref{eq38}\;
 			}
 		}
 		
	}
	
\end{algorithm}

Then the algorithm runs to the next time slot. Vehicles' positions, channels and number changes, we can get each vehicle's next state ${{s}_{{{c}_{i}}}}(t+1)$. Each vehicle stores the transition tuple $\left( {{s}_{{{c}_{i}}}}(t),{{p}_{{{c}_{i}}}}(t),{{r}_{{{c}_{i}}}}(t),{{s}_{{{c}_{i}}}}(t+1) \right)$ in its own experience replay buffer ${{\mathcal{B}}_{{{c}_{i}}}}$. If the data number $|{{\mathcal{B}}_{{{c}_{i}}}}|$ in ${{\mathcal{B}}_{{{c}_{i}}}}$ exceeds ${{\mathcal{I}}_{{{c}_{i}}}}$, vehicle ${{c}_{i}}$ begins ${{I}_{{{c}_{i}}}}$ iterations of model training and updating. A new road graph $G(t)$ also can be obtained based on the current road scenario and vehicles' model training. The newly $G(t)$ is inputted to the GNN network to generate $\bar{\psi }\left( t \right)$. Then the GNN transition tuple $\left( G(t),\bar{\psi }\left( t \right),\bar{\phi }(t),G(t+1) \right)$is stored in the GNN experience replay buffer ${{\mathcal{B}}_{g}}$. 
For each iteration of vehicle's SAC model training, a batch of training data is constructed by randomly selecting $\mathcal{M}$ tuples from ${{\mathcal{B}}_{{{c}_{i}}}}$. Let $\left( {{s}_{{{c}_{i}},i}},{{a}_{{{c}_{i}},i}},{{r}_{{{c}_{i}},i}},{{{{s}'}}_{{{c}_{i}},i}} \right)(i=1,2,\cdots ,\mathcal{M})$ be the $i$th tuple in the mini-batch for vehicle ${{c}_{i}}$. For each tuple $i$, ${{s}_{{{c}_{i}},i}}$ is inputted into the actor network, producing the action ${{\tilde{a}}_{{{c}_{i}},i}}$. The gradient of the loss function for ${{\beta }_{{{c}_{i}}}}$ is calculated as
\begin{equation}
\begin{aligned}
&{{\nabla }_{{{\beta }_{{{c}_{i}}}}}}{{J}_{{{c}_{i}}}}\left( {{\beta }_{{{c}_{i}}}} \right)= \\
&{{\nabla }_{{{\beta }_{{{c}_{i}}}}}}{{E}_{{{a}_{{{c}_{i}},i}}\sim{{\pi }_{{{c}_{i}},{{\theta }_{{{c}_{i}}}}}}}}\left[ -{{\beta }_{{{c}_{i}}}}\log{{\pi }_{{{c}_{i}},{{\theta }_{{{c}_{i}}}}}}\left( {{{\tilde{a}}}_{{{c}_{i}},i}}|{{s}_{{{c}_{i}},i}} \right)-{{\beta }_{{{c}_{i}}}}{{\overline{H}}_{{{c}_{i}}}} \right]	
\end{aligned}
\label{eq29}.
\end{equation}
Next ${{s}_{{{c}_{i}},i}}$ and ${{\tilde{a}}_{{{c}_{i}},i}}$ are inputted into the two critic networks to obtain the action-value functions ${{Q}_{\varphi _{1}^{{{c}_{i}}}}}\left( {{s}_{{{c}_{i}},i}},{{{\tilde{a}}}_{{{c}_{i}},i}} \right)$ and ${{Q}_{\varphi _{2}^{{{c}_{i}}}}}\left( {{s}_{{{c}_{i}},i}},{{{\tilde{a}}}_{{{c}_{i}},i}} \right)$. Then the gradient of the loss function for actor network parameters ${{\theta }_{{{c}_{i}}}}$ is calculated as
\begin{equation}
\begin{aligned}
  & {{\nabla }_{{{\theta }_{{{c}_{i}}}}}}{{J}_{{{c}_{i}}}}({{\theta }_{{{c}_{i}}}})= \\
  & {{\nabla }_{{{\theta }_{{{c}_{i}}}}}}{{\beta }_{{{c}_{i}}}}\log \left( {{\pi }_{{{\theta }_{{{c}_{i}}}}}}\left( {{{\tilde{a}}}_{{{c}_{i}},i}}\left| {{s}_{{{c}_{i}},i}} \right. \right) \right)+{{\nabla }_{{{\theta }_{{{c}_{i}}}}}}f\left( {{\varepsilon }_{{{c}_{i}}}};{{s}_{{{c}_{i}},i}} \right)\cdot  \\ 
 & \left( {{\nabla }_{{{a}_{{{c}_{i}},i{ }}}}}{{\beta }_{{{c}_{i}}}}\log \left( {{\pi }_{\phi }}\left( {{{\tilde{a}}}_{{{c}_{i}},i}}\left| {{s}_{{{c}_{i}},i}} \right. \right) \right)-{{\nabla }_{{{{\tilde{a}}}_{{{c}_{i}},i}}}}{{Q}_{{{c}_{i}}}}\left( {{s}_{{{c}_{i}},i}},{{{\tilde{a}}}_{{{c}_{i}},i}} \right) \right)  
\end{aligned}
\label{eq30},
\end{equation}
where ${{\varepsilon }_{{{c}_{i}}}}$ is noise sampled from a multivariate normal distribution and $f\left( {{\varepsilon }_{{{c}_{i}}}};{{s}_{{{c}_{i}},i}} \right)$ is a reparameterization trick function for ${{\tilde{a}}_{{{c}_{i}},i}}$. ${{Q}_{{{c}_{i}}}}\left( {{s}_{{{c}_{i}},i}},{{{\tilde{a}}}_{{{c}_{i}},i}} \right)$ is calculated as $min\left\{ {{Q}_{\varphi _{1}^{{{c}_{i}}}}}\left( {{s}_{{{c}_{i}},i}},{{{\tilde{a}}}_{{{c}_{i}},i}} \right),{{Q}_{\varphi _{2}^{{{c}_{i}}}}}\left( {{s}_{{{c}_{i}},i}},{{{\tilde{a}}}_{{{c}_{i}},i}} \right) \right\}$.
Next, the algorithm computes the gradients for the two critic network parameters $\varphi _{1}^{{{c}_{i}}}$ and $\varphi _{2}^{{{c}_{i}}}$. For each tuple $i$ in the mini-batch, the states ${{s}_{{{c}_{i}},i}}$ and actions ${{a}_{{{c}_{i}},i}}$ are inputted into the two critic networks, producing the action-value functions ${{Q}_{\varphi _{1}^{{{c}_{i}}}}}\left( {{s}_{{{c}_{i}},i}},{{a}_{{{c}_{i}},i}} \right)$ and ${{Q}_{\varphi _{2}^{{{c}_{i}}}}}\left( {{s}_{{{c}_{i}},i}},{{a}_{{{c}_{i}},i}} \right)$. Additionally, the next state ${{{s}'}_{{{c}_{i}},i}}$ is inputted into the actor network to output ${{{a}'}_{{{c}_{i}}}}$, which is then fed into the two target critic networks to output ${{Q}_{\bar{\varphi }_{1}^{{{c}_{i}}}}}\left( {{{{s}'}}_{{{c}_{i}},i}},{{{{a}'}}_{{{c}_{i}}}} \right)$ and ${{Q}_{\bar{\varphi }_{2}^{{{c}_{i}}}}}\left( {{{{s}'}}_{{{c}_{i}},i}},{{{{a}'}}_{{{c}_{i}}}} \right)$. The target action-value is then calculated as
\begin{equation}
\begin{aligned}
& {{\hat{Q}}_{{{c}_{i}}}}\left( {{{{s}'}}_{{{c}_{i}},i}},{{{{a}'}}_{{{c}_{i}}}} \right) =-{{\beta }_{{{c}_{i}}}}\log \left( {{\pi }_{{{\theta }_{{{c}_{i}}}}}}\left( {{{{a}'}}_{{{c}_{i}}}}\left| {{{{s}'}}_{{{c}_{i}},i}} \right. \right) \right) \\
& +min\left\{ {{Q}_{\bar{\varphi }_{1}^{{{c}_{i}}}}}\left( {{{{s}'}}_{{{c}_{i}},i}},{{{{a}'}}_{{{c}_{i}}}} \right),{{Q}_{\bar{\varphi }_{2}^{{{c}_{i}}}}}\left( {{{{s}'}}_{{{c}_{i}},i}},{{{{a}'}}_{{{c}_{i}}}} \right) \right\}{ } 
\end{aligned}
\label{eq31},	
\end{equation}
The gradients for the loss functions of $\varphi _{1}^{{{c}_{i}}}$ and $\varphi _{2}^{{{c}_{i}}}$ are calculated as
\begin{equation}
\begin{aligned}
& {{\nabla }_{\varphi _{m}^{{{c}_{i}}}}}{{J}_{{{c}_{i}}}}\left( \varphi _{m}^{{{c}_{i}}} \right){ }={{\nabla }_{\varphi _{m}^{{{c}_{i}}}}}{{Q}_{\varphi _{m}^{{{c}_{i}}}}}\left( {{s}_{{{c}_{i}},i}},{{a}_{{{c}_{i}},i}} \right)\cdot \\
& \left( {{Q}_{\varphi _{m}^{{{c}_{i}}}}}\left( {{s}_{{{c}_{i}},i}},{{a}_{{{c}_{i}},i}} \right)-{{r}_{{{c}_{i}},i}}+\gamma {{{\hat{Q}}}_{{{c}_{i}}}}\left( {{{{s}'}}_{{{c}_{i}},i}},{{{{a}'}}_{{{c}_{i}}}} \right) \right), m \in \left\{ 1,2\right\}  
\end{aligned}
\label{eq32},	
\end{equation} 

Using the Adam optimizer and based on the gradients ${{\nabla }_{{{\beta }_{{{c}_{i}}}}}}{{J}_{{{c}_{i}}}}\left( {{\beta }_{{{c}_{i}}}} \right)$, ${{\nabla }_{{{\theta }_{{{c}_{i}}}}}}{{J}_{{{c}_{i}}}}({{\theta }_{{{c}_{i}}}})$, ${{\nabla }_{\varphi _{1}^{{{c}_{i}}}}}{{J}_{{{c}_{i}}}}\left( \varphi _{1}^{{{c}_{i}}} \right)$, and ${{\nabla }_{\varphi _{2}^{{{c}_{i}}}}}{{J}_{{{c}_{i}}}}\left( \varphi _{2}^{{{c}_{i}}} \right)$, the parameters ${{\beta }_{{{c}_{i}}}}$, ${{\theta }_{{{c}_{i}}}}$, $\varphi _{1}^{{{c}_{i}}}$, and $\varphi _{2}^{{{c}_{i}}}$ are updated through gradient descent. Note that after every ${{\tilde{I}}_{{{c}_{i}}}}$ iterations of training, the parameters of the two target critic networks are updated as
\begin{equation}
\begin{aligned}
\bar{\varphi }_{m}^{{{c}_{i}}}:={{\tau }_{m}}\varphi _{m}^{{{c}_{i}}}+\left( 1-{{\tau }_{m}} \right)\bar{\varphi }_{m}^{{{c}_{i}}}, m \in \left\{ 1,2\right\} 
\end{aligned}
\label{eq34},	
\end{equation}
where ${{\tau }_{m}}$ is constant satisfying ${{\tau }_{m}}\ll 1$. 

As for GNN, when the buffer ${{\mathcal{B}}_{g}}$ contains data number $|{{\mathcal{B}}_{g}}|$ exceeding ${{\mathcal{I}}_{g}}$, the algorithm trains and updates the GNN network, GNN critic network, and target GNN critic network every ${{T}_{g}}$ time slots for ${{I}_{g}}$ iterations. The RSU randomly selects ${{\mathcal{M}}_{g}}$ tuples from  ${{\mathcal{B}}_{g}}$ to form a training batch. Let $\left( {{G}_{i}},{{{\bar{\psi }}}_{i}},{{{\bar{\phi }}}_{i}},{{{{G}'}}_{i}} \right)(i=1,2,\cdots ,{{\mathcal{M}}_{g}})$ be the $i$th tuple in the RSU's mini-batch. The loss function for the GNN network model parameters ${{\theta }_{g}}$ is defined as
\begin{equation}
{{\mathcal{L}}_{{{\theta }_{g}}}}=-\frac{1}{{{\mathcal{M}}_{g}}}\sum\limits_{i=1}^{{{\mathcal{M}}_{g}}}{{{Q}_{{{\varphi }_{g}}}}}({{G}_{i}},{{\bar{\psi }}_{i}})
\label{eq36},	
\end{equation}
where ${{Q}_{{{\varphi }_{g}}}}({{G}_{i}},{{\bar{\psi }}_{i}})$ represents the feature value function, i.e., the GNN critic network evaluates the quality of generated feature values of GNN network. The loss function for the GNN critic network model parameters ${{\varphi }_{g}}$ is defined as
\begin{equation}
{{\mathcal{L}}_{{{\varphi }_{g}}}}=-\frac{1}{{{\mathcal{M}}_{g}}}\sum\limits_{i=1}^{{{\mathcal{M}}_{g}}}{{{\left[ {{Q}_{{{\varphi }_{g}}}}({{G}_{i}},{{{\bar{\psi }}}_{i}})-{{{\bar{\phi }}}_{i}}-\gamma \cdot {{Q}_{{{{\bar{\varphi }}}_{g}}}}({{{{G}'}}_{i}},{{{{\bar{\psi }}'}}_{i}}) \right]}^{2}}}
\label{eq37},	
\end{equation}
where ${{{\bar{\psi }}'}_{i}}$ represents the node feature vector obtained by inputting ${{{G}'}_{i}}$ into the GNN network. After every ${{\tilde{I}}_{g}}$ iterations of training, the GNN critic network is updated as
\begin{equation}
{{\bar{\varphi }}_{g}}:={{\tau }_{g}}{{\varphi }_{g}}+\left( 1-{{\tau }_{g}} \right){{\bar{\varphi }}_{g}}
\label{eq38},	
\end{equation}
where ${{\tau }_{g}}$ is a constant satisfying ${{\tau }_{g}}\ll 1$. Finally, the algorithm runs to the next time slot. When the algorithm executes to $R^{\max}$, it indicates that the training has ended.

\paragraph{Testing Stage}
The testing stage omits the critic network, target critic network, GNN network, GNN critic network and the target GNN critic network. During testing stage, the optimal strategy is evaluated using the optimized parameters of the actor network ${{(\theta _{a}^{global})}^{*}}$. 
\section{Numerical Simulation and Analysis}
In this section, we evaluate the performance of our proposed FGNN-MADRL scheme through simulation experiments and discuss the results obtained. The simulation experiments are implemented using Python 3.7, and the simulation scenario is constructed based on the system model. Table \ref{tab2} lists the parameters used in the simulation environment. 
\begin{table}[t]
	\caption{Environment parameters in the simulation.}
	\label{tab2}
	\footnotesize
	\centering
	\begin{tabular}{|c|c|c|c|}
		\hline
		\textbf{Parameter} &\textbf{Value} &\textbf{Parameter} &\textbf{Value}\\
		$L$ & 4 & $\tau$ & 0.02 s \\ 
		${{p}_{\max}}$ & 20 Watts & ${{D}_{r}}$ & 250 m \\ 
		${{d}_{\min}}$ & 0.1 MB & ${{d}_{\max}}$ & 10 MB \\ 
		$\mu$ & 0.2 s & ${{{D}}_{c}}$ & 100 m \\ 
		$\delta$ & 0.9999 & ${{\omega}_{{2}}}$ & 0.9999 \\ 
		${{R}^{\max}}$ & 20000 s & ${{R}^{test}}$ & 2000 s \\ 
		$B$ & 200 MHz & ${{\sigma}^{2}}$ & $3.98\times {{10}^{-14}}$ Watts \\ 
		${{d}_{{cor}}}$ & 10 & ${{f}_{c}}$ & $28\times {{10}^{9}}$ Hz \\ 
		$c$ & $3\times {{10}^{8}}$ m/s & ${{\sigma}_{s}}$ & 2.2 dB \\ 
		\hline
	\end{tabular}
\end{table}
\begin{table}
	\caption{SAC Hyperparameters}
	\label{tab3}
	\footnotesize
	\centering
	\begin{tabular}{|c|c|c|c|}
		\hline
		\textbf{Parameter} &\textbf{Value} &\textbf{Parameter} &\textbf{Value}\\
		\hline
		${{\alpha}^{A}}$ & ${{10}^{-4}}$ & ${{\alpha}^{C}}$ & ${{10}^{-3}}$ \\ 
		${{D}_{s}}$ & 500 & ${{\mathcal{I}}_{{{c}_{i}}}}$ & 256 \\ 
		$\mathcal{M}$ & 128 & ${{\tilde{I}}_{{{c}_{i}}}}$ & 1 \\ 
		${{\tau}_{1}}$ & 0.005 & ${{\tau}_{2}}$ & 0.005 \\ 
		Reward Scaling Factor & 0.1 & Activation Function & ReLU \\
		\hline
		\end{tabular}
\end{table}
Both actor and critic networks in SAC use four-layer fully connected DNNs, with two middle hidden layers each containing 256 neurons. We consider the heterogeneity of each vehicle, meaning each vehicle has a different number of SAC model training iterations ${{I}_{{{c}_{i}}}}=\{5,10,20,40,50\}$. Each generated vehicle randomly selects a value from ${{I}_{{{c}_{i}}}}$ as its iteration number. In addition, some other hyperparameters are adapted from \cite{ref31}. Table \ref{tab3} lists the remaining hyperparameters for the SAC network. ${{\alpha }^{A}}$ and ${{\alpha }^{C}}$ are the learning rates for the actor and critic networks, respectively. 
\begin{table}
	\caption{GNN and GNN Critic Hyperparameters}
	\label{tab4}
	\footnotesize
	\centering
	\begin{tabular}{|c|c|c|c|}
		\hline
		\textbf{Parameter} &\textbf{Value} &\textbf{Parameter} &\textbf{Value}\\
		\hline
		${{D}_{g}}$ & 5000 & ${{\mathcal{M}}_{g}}$ & 128 \\ 
		${{\mathcal{I}}_{g}}$ & 256 & ${{I}_{{{c}_{i}}}}$ & 5 \\ 
		${{\alpha}^{G}}$ & 0.001 & ${{\alpha}^{GC}}$ & 0.001 \\ 
		Optimizer & Adam Optimizer & ${{\tilde{I}}_{g}}$ & 1 
		\\ \hline
		\end{tabular}
\end{table}

The GNN network uses a four-layer DNN, with the neuron numbers in the two middle hidden DNN layers being 128 and 64, respectively. The activation function for the hidden layers is the Tanh function. Both the GNN critic network and the target GNN critic network also use four-layer DNNs, with 256 neurons in each of the two middle hidden layers. The hyperparameters for the GNN critic network are basically the same as those in the SAC model's critic network. Table \ref{tab4} lists the hyperparameters used in the GNN network and GNN critic network. ${{\alpha }^{G}}$ and ${{\alpha }^{GC}}$ are the learning rates for the GNN critic network and the target GNN critic network, respectively.

\subsection{Training Stage}

\begin{figure}[h!]
	\center
	\includegraphics[scale = 0.45]{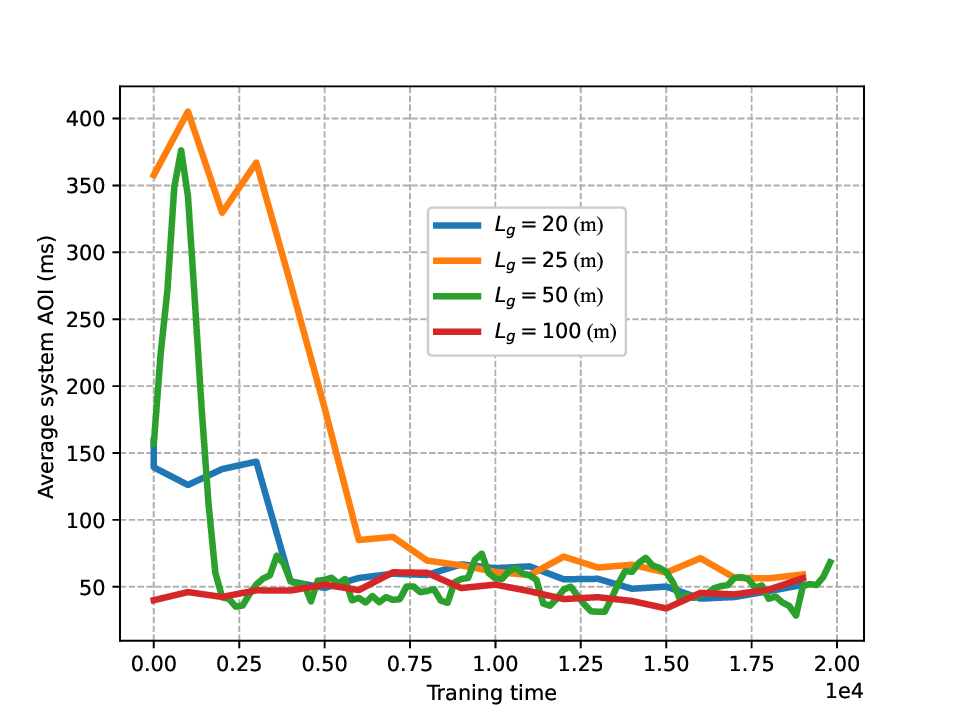}
	\caption{Learning curve}
	\label{fig_training}
\end{figure}

Fig. \ref{fig_training} displays the learning curves of the SAC model over time for different road segment lengths ${{L}_{g}}$. A larger ${{L}_{g}}$ indicates fewer nodes in the GNN. As seen in the figure, all learning curves gradually decline over time and eventually stabilize, indicating that all models can converge. It is also observed that as ${{L}_{g}}$ increases, the model converges more quickly. This is because a smaller road segment length ${{L}_{g}}$ leads to more nodes in the vehicle graph structure, making the GNN network more complex. Therefore, more time is required to train the GNN network, which in turn increases the training time of the SAC model. 
The convergence rate for ${{L}_{g}}$ equal to 20 m is actually faster than for ${{L}_{g}}$ equal to 25 m. 
This is because, when ${{L}_{g}}$ is 20 m, there are more edges in the graph, which paradoxically facilitates the training of the model. 
\subsection{Testing Stage}
To ensure more accurate test results, all simulation outcomes during the testing stage are averaged after 50 experiments. To validate the effectiveness of our proposed FGNN-MADRL scheme, we compare it with the following three algorithms:
\begin{itemize}
\item{Global Federated Multi-Agent Reinforcement Learning (GFSAC): In this method, agents do not perform local aggregation. Once a vehicle completes its local model training, it uploads the model directly to the RSU for global federated averaging aggregation.}

\item{Local Federated Multi-Agent Reinforcement Learning (LFSAC): In this method, agents perform local aggregation. After completing local model training, vehicles first average aggregate locally with other vehicles within their communication range. This method does not use GNN to generate model aggregation weights. Vehicles upload their model to the RSU for global aggregation just before leaving the RSU coverage area.}

\item{ Game-based Dynamic Best Response for Cooperative Vehicle Task Offloading (GDBR) \cite{ref6}: This method defines the global AOI as the utility function of the game. It considers the best response probability of other vehicles offloading in the previous time slot as the price function of the game. The method iteratively updates the best response probability for vehicle offloading tasks based on the utility and price functions, eventually converting the best response probability into the transmission power for the vehicle's offloading tasks.}
\end{itemize}

\begin{figure}[h]
  \centering
  \subfigure[Average system AoI]{
    \begin{minipage}{7.5cm}
    \includegraphics[scale=0.45]{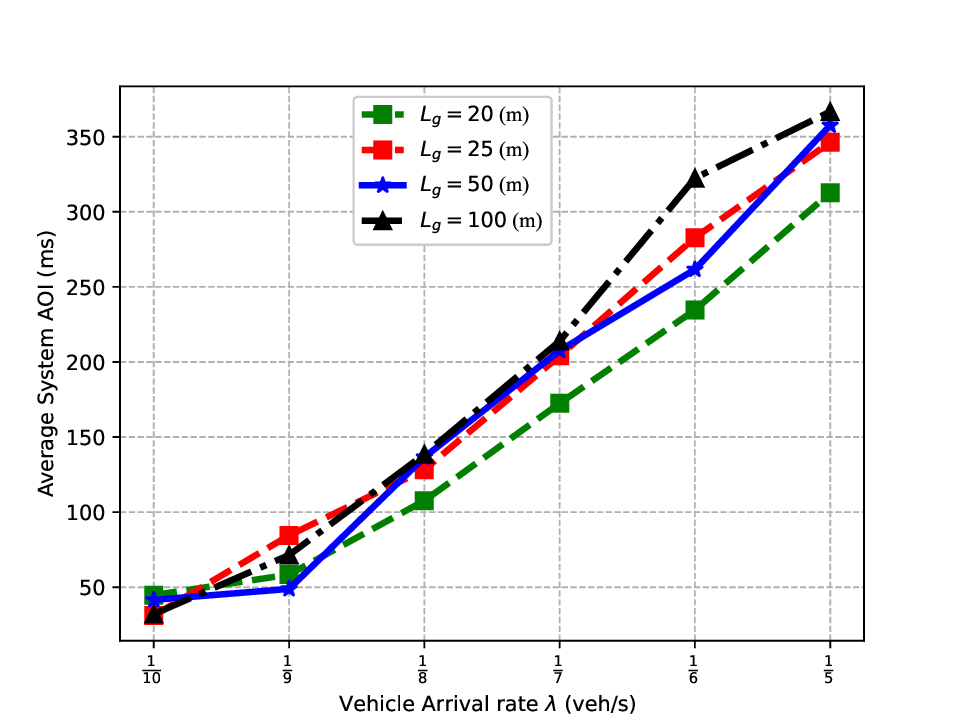}
    \end{minipage}
    \label{fig_gnnsize_density_aoi}
    }
  \subfigure[Average system power]{
    \begin{minipage}{7.5cm}
    \includegraphics[scale=0.45]{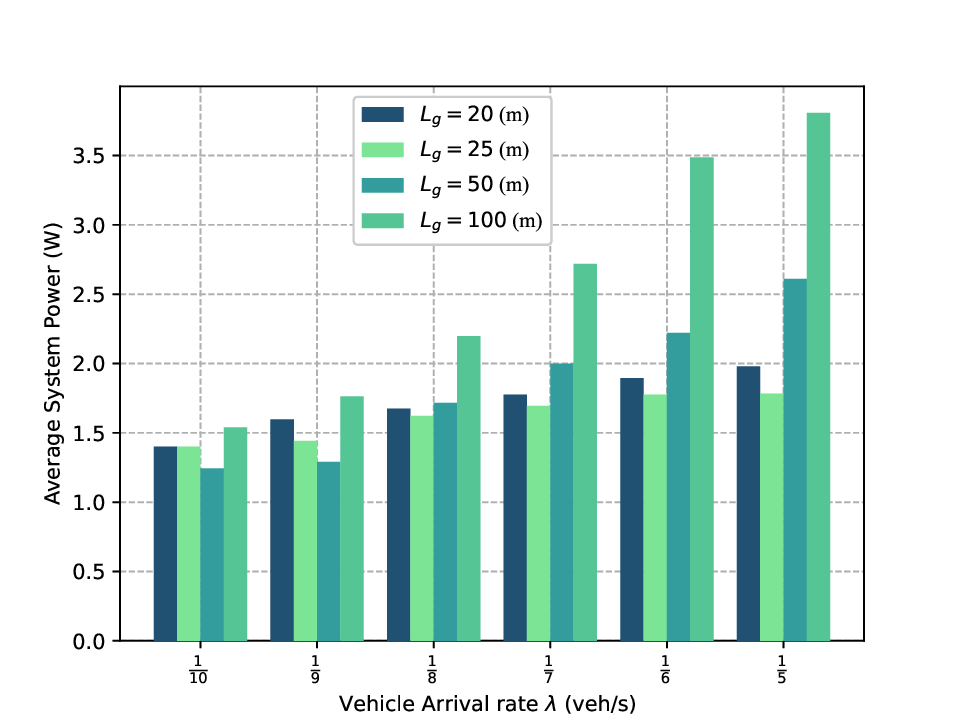}
    \end{minipage}
    \label{fig_gnnsize_density_power}
    }
    \caption{The performance of different ${{L}_{g}}$ trained models under various $\lambda$.}
\end{figure}

Fig. \ref{fig_gnnsize_density_aoi}-\ref{fig_gnnsize_density_power} show the average AOI and power of all vehicles under different vehicle arrival rates and ${{L}_{g}}$. In this experiment, the speed of all vehicles is set to 30 Km/h. From Fig. \ref{fig_gnnsize_density_aoi}, it is observed that the system's average AOI increases with the increase in vehicle density. This is because, as the number of vehicles increases, the communication interference between them also increases, leading to a reduction in the vehicles' transmission rate. It is also evident from Fig. \ref{fig_gnnsize_density_aoi} that the average AOI increases with the increase in ${{L}_{g}}$. This is because a larger ${{L}_{g}}$ results in fewer nodes in the vehicle graph structure, making the graph simpler and thus less effective at extracting vehicular feature information. Consequently, the model weights produced are less reflective of vehicle information, reducing the accuracy of model training.
In Fig. \ref{fig_gnnsize_density_power}, as the vehicle arrival rate increases, the power consumption of the solutions under different ${{L}_{g}}$ trained models also increases. This is due to the increased interference among vehicles as their number grows, requiring more power for task offloading in order to reduce AoI. It is also noticeable that as ${{L}_{g}}$ increases, the average power consumption also increases. This is because a larger ${{L}_{g}}$ leads to less information being extracted by the GNN, thus hindering the training of more effective model performance.

\begin{figure}[h]
  \centering
  \subfigure[Average system AoI]{
    \begin{minipage}{7.5cm}
    \includegraphics[scale=0.45]{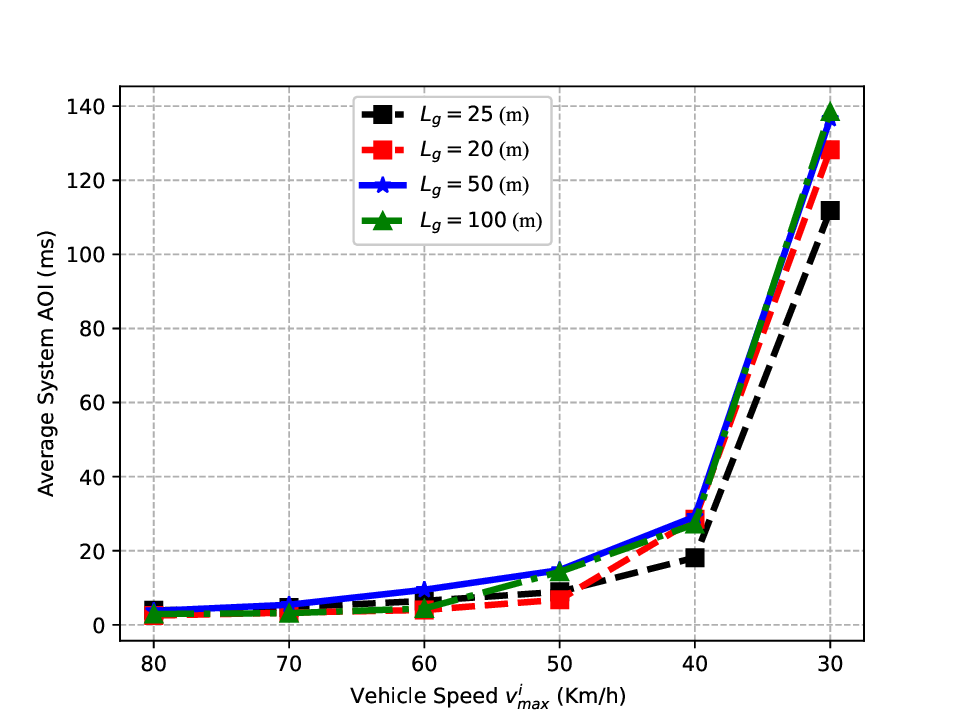}
    \end{minipage}
    \label{fig_gnnsize_speed_aoi}
    }
  \subfigure[Average system power]{
    \begin{minipage}{7.5cm}
    \includegraphics[scale=0.45]{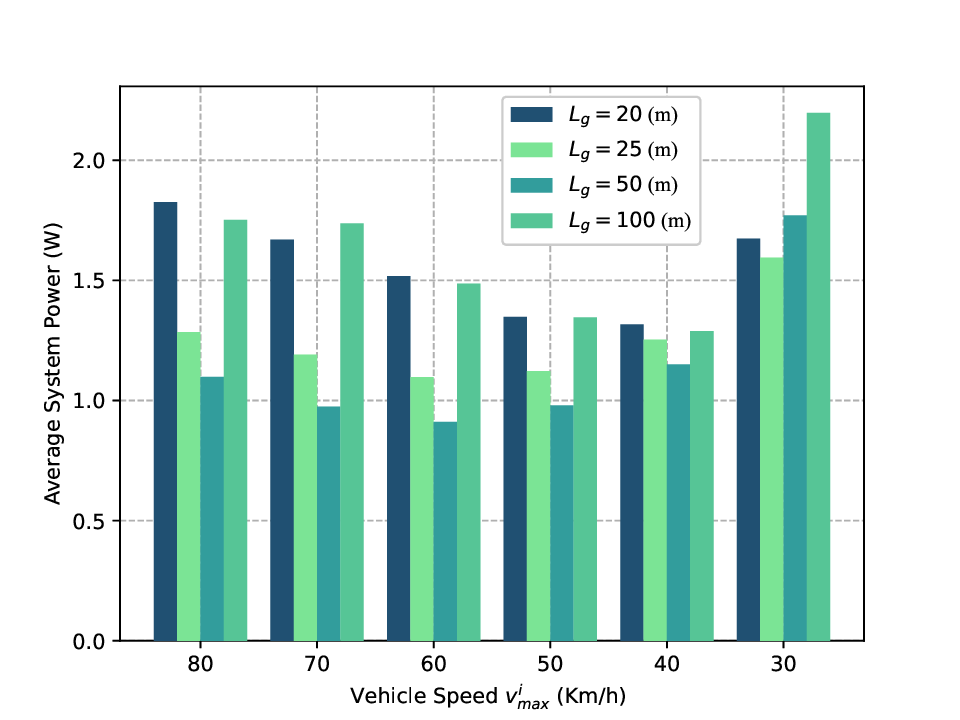}
    \end{minipage}
    \label{fig_gnnsize_speed_power}
    }
    \caption{The performance of different ${{L}_{g}}$  trained models under various $v^{i}_{\max}$.}
\end{figure}

Fig. \ref{fig_gnnsize_speed_aoi}-\ref{fig_gnnsize_speed_power} show the average AoI and power of all vehicles under different ${{L}_{g}}$ training models at various vehicle speeds. In this experiment, the vehicle arrival rate is set to \(\frac{1}{8}\) vehicles per second. It is also evident from Fig. \ref{fig_gnnsize_speed_aoi}, the AoI for all four different models increases as vehicle speed decreases. This is due to the increase in the number of vehicles on the road and the resulting increase in interference between vehicles as their speed decreases. Furthermore, when ${{L}_{g}}$ is greater than 25m, the average AoI increases with ${{L}_{g}}$, which can be attributed to the deterioration in the quality of model training. However, when ${{L}_{g}}$ is equal to 20m, the performance in terms of average AoI is worse than when ${{L}_{g}}$ is 25m. This is because at ${{L}_{g}}$ equal to 20m, vehicles spend a very short time at nodes, preventing the GNN from effectively extracting vehicle information. This issue is more pronounced at lower vehicle arrival rates and higher speeds.

It can be seen from Fig. \ref{fig_gnnsize_speed_power}, as vehicle speed decreases, the power usage of the four models initially decreases and then increases. This is because, with decreasing speed, vehicles spend more time at nodes, allowing the GNN to more effectively extract vehicle information for training. However, when the vehicle speed is 30 Km/h, due to the very low speed, there is an increase in the number of vehicles, which in turn increases interference. To reduce AoI, vehicles increase power to compete for channel resources for task offloading.

\begin{figure}[t!]
  \centering
  \subfigure[Average system AoI]{
    \begin{minipage}{7.5cm}
    \includegraphics[scale=0.45]{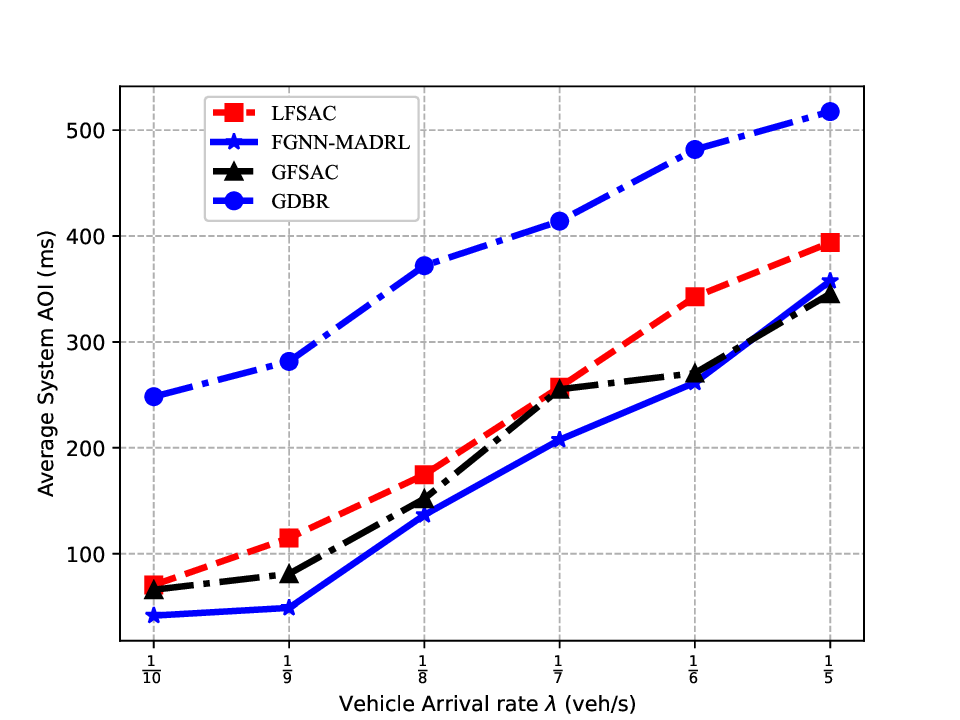}
    \end{minipage}
    \label{fig_gnnsize_compare_density_aoi}
    }
  \subfigure[Average system power]{
    \begin{minipage}{7.5cm}
    \includegraphics[scale=0.45]{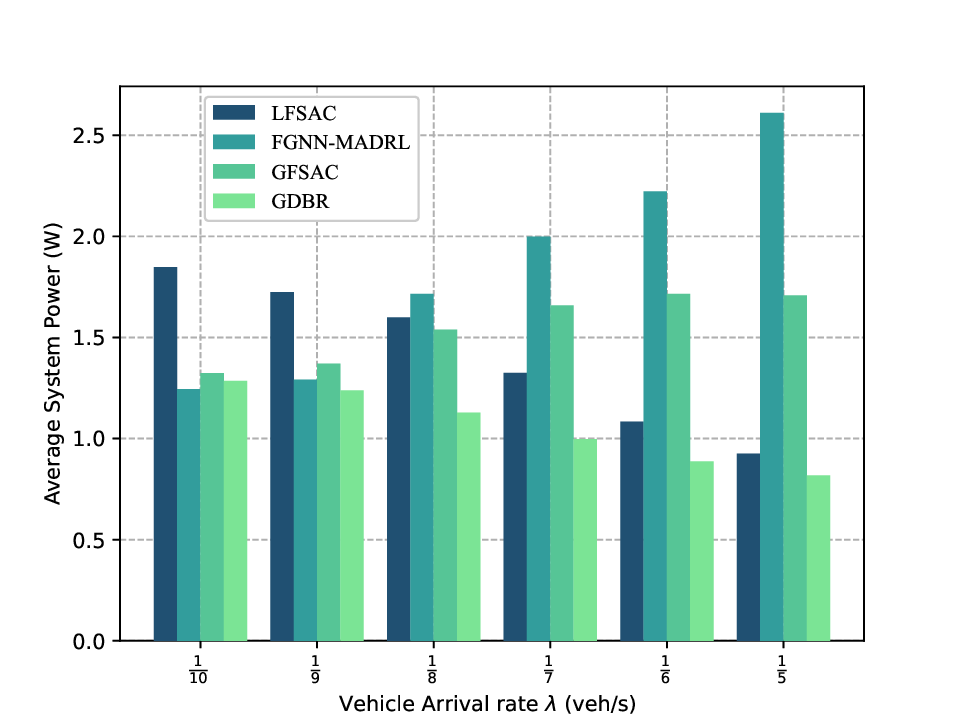}
    \end{minipage}
    \label{fig_gnnsize_compare_density_power}
    }
    \subfigure[Average system thoughout]{
    \begin{minipage}{7.5cm}
    \includegraphics[scale=0.45]{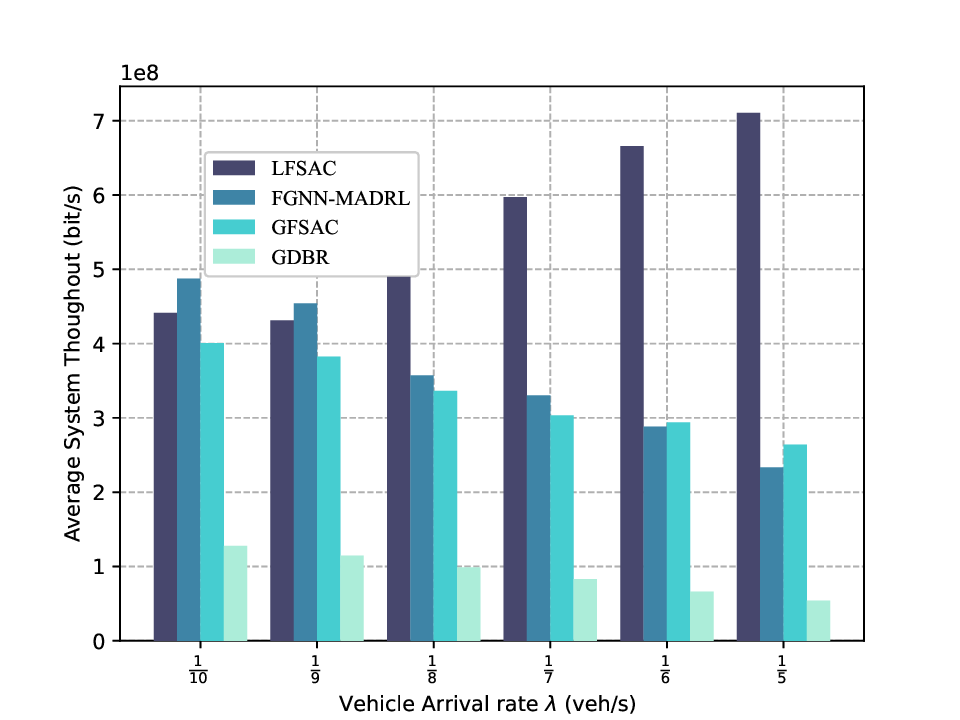}
    \end{minipage}
    \label{fig_gnnsize_compare_density_thoughout}
    }
    \caption{The performance of different schemes under various $\lambda$.}
\end{figure}
Fig. \ref{fig_gnnsize_compare_density_aoi}-\ref{fig_gnnsize_compare_density_thoughout} present the average AOI, power and throughput of all vehicles under different schemes under various $\lambda$. In this experiment, the speed of all vehicles is set to 30 Km/h. Our FGNN-MADRL scheme is tested with ${{L}_{g}}=50m$. From Fig. \ref{fig_gnnsize_compare_density_aoi}, it can be observed that as $\lambda$ increases, our FGNN-MADRL scheme has the smallest average AOI, demonstrating its superiority. The performance of GDBR is the worst, as it makes decisions based on probabilities. The performances of LFSAC and GFSAC are better than GDBR, as these two schemes utilize RL methods, allowing some degree of cooperative offloading between vehicles. However, they perform worse than our FGNN-MADRL scheme because their model training involves only average federated aggregation, lacking personalized features in the RL model.
It can be seen from Fig. \ref{fig_gnnsize_compare_density_power}, with the increase in $\lambda$, the average power for both FGNN-MADRL and GFSAC also increases. This is due to the greater interference among an increasing number of vehicles, necessitating more power for transmission. The average power of FGNN-MADRL is higher than that of GFSAC, as it allocates more power to achieve better average AOI performance as shown in Fig. \ref{fig_gnnsize_compare_density_aoi}. The power consumption for LFSAC and GBDR does not increase with the rising vehicle arrival rate, indicating their inability to adapt to scenarios with high vehicle density.
From Fig. \ref{fig_gnnsize_compare_density_thoughout}, the average throughput for FGNN-MADRL, GFSAC, and GBDR decreases as $\lambda$ increases. This is because the increase in vehicle numbers leads to more interference, thus reducing throughput. On the other hand, the average throughput for LFSAC increases with the rising $\lambda$, due to that vehicles do not offload tasks in a timely manner.

\begin{figure}[t!]
  \centering
  \subfigure[Average system AoI]{
    \begin{minipage}{7.5cm}
    \includegraphics[scale=0.45]{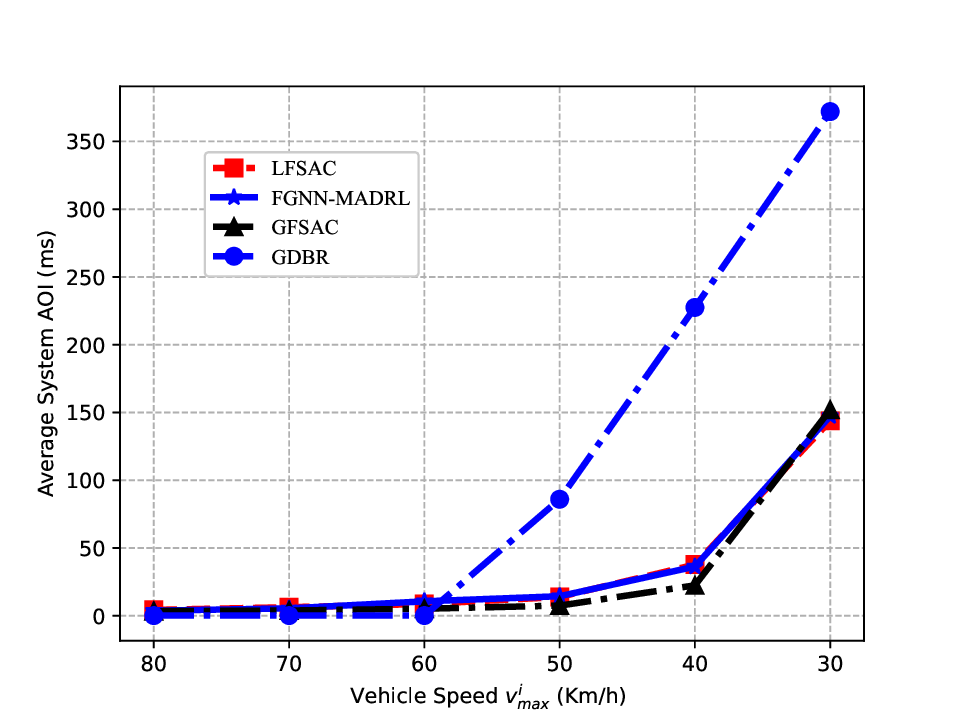}
    \end{minipage}
    \label{fig_gnnsize_compare_speed_aoi}
    }
  
  \subfigure[Average system power]{
    \begin{minipage}{7.5cm}
    \includegraphics[scale=0.45]{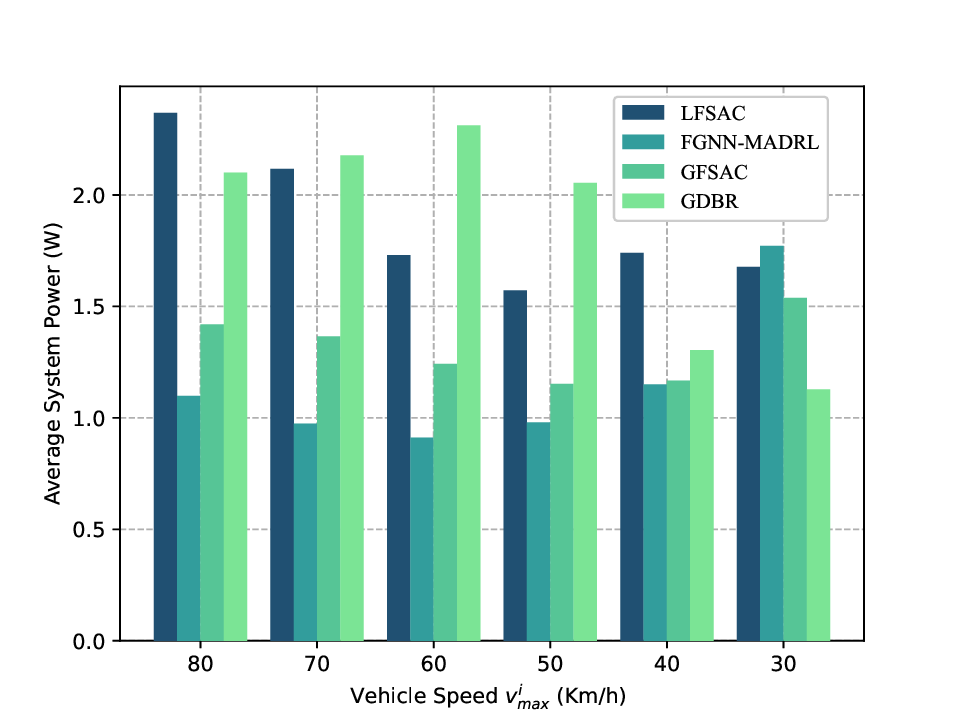}
    \end{minipage}
    \label{fig_gnnsize_compare_speed_power}
    }
    \subfigure[Average system thoughout]{
    \begin{minipage}{7.5cm}
    \includegraphics[scale=0.45]{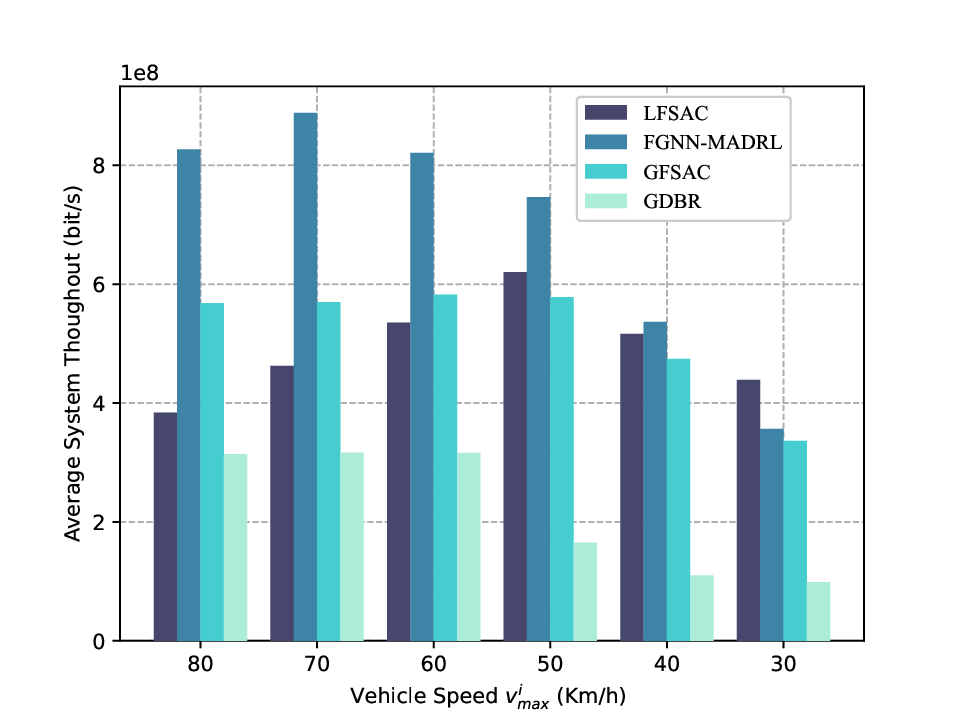}
    \end{minipage}
    \label{fig_gnnsize_compare_speed_thoughout}
    }
    \caption{The performance of different schemes under various $v^{i}_{\max}$.}
\end{figure}
Fig. \ref{fig_gnnsize_compare_speed_aoi}-\ref{fig_gnnsize_compare_speed_thoughout} display the average AOI, power, and throughput of all vehicles under different algorithms at varying vehicle speeds. In this experiment, the vehicle arrival rate is set to \(\frac{1}{8}\) vehicles per second. Our FGNN-MADRL scheme is tested with ${{L}_{g}}=50m$. It can be seen from Fig. \ref{fig_gnnsize_compare_speed_aoi}, the AOI for all four methods increases as vehicle speed decreases, due to the increase in the number of vehicles on the road and the resultant increase in interference between vehicles. FGNN-MADRL, LFSAC, and GFSAC all exhibit good AOI performance, as these three schemes utilize RL methods to make appropriate decisions. GDBR shows the worst performance in terms of AOI because it makes decisions based on probability.

In Fig. \ref{fig_gnnsize_compare_speed_power}, as vehicle speed decreases, the average power of FGNN-MADRL gradually increases. This is because the decrease in vehicle speed leads to an increase in the number of vehicles and thus increased interference. To reduce the AOI, higher transmission power is needed. Additionally, FGNN-MADRL uses the least average power, indicating that it can achieve better AOI performance with less power, demonstrating the superiority of our scheme. The other three schemes do not show a consistent trend of change in power with the reduction in vehicle speed, as they do not extract features.

In Fig. \ref{fig_gnnsize_compare_speed_thoughout}, the average throughput of FGNN-MADRL and GDBR gradually decreases as vehicle speed decreases, due to the increased number of vehicles and interference. GDBR has the lowest average throughput because it allocates power based on probability. LFSAC and GFSAC do not show a consistent trend of increase or decrease in throughput with the reduction in vehicle speed, as they do not extract the vehicle's road graph structure and thus cannot adapt to changes in vehicle speed. FGNN-MADRL has the highest average throughput because it uses RL methods for cooperative decision-making, reducing interference between vehicles and making reasonable cooperative allocations based on the current environment, thereby proving the superiority of our scheme.
\section{Conclusions}
In this paper, we addressed the problem of optimizing AoI in a multi-vehicle scenario. We proposed an innovative FGNN-MADRL algorithm, which integrates GNN with MADRL to optimize AoI. The key characteristic of our model is that road scenarios is first modeled as a graph and an effective FL framework that combines both distributed based on GNN and centralized federated aggregation is employed. Additionally, we introduced a MADRL algorithm designed to reduce decision complexity. Conclusions are drawn as follows: 
\begin{itemize}
\item{The structure of a GNN impacts the training of models. Both an excess or a deficiency of GNN nodes can hinder the effective training of DRL. This is because more GNN nodes mean shorter vehicle dwell times at each node, while fewer GNN nodes lead to a simpler network structure with weaker information extraction capabilities.}
\item{Compared to other distributed federated algorithms, FGNN-MADRL can extract information about road vehicles, such as vehicle density, speed, and the status of model training. As a result, FGNN-MADRL adapts well to dynamic scenarios and effectively reduces AoI.}
\item{In contrast to other non-RL algorithms, FGNN-MADRL facilitates collaboration among vehicles, thereby reducing the average age of information. This occurs because vehicles can observe local conditions and make sensible decisions. Training models with the assistance of a GNN takes into account information from other vehicles.}
\end{itemize}

\bibliographystyle{IEEEtran}
\bibliography{reference}

\end{document}